\documentclass[lettersize,journal]{IEEEtran}  
\usepackage{graphicx} 
\usepackage[hyphens]{url}  
\usepackage{algorithm}
\usepackage{algorithmic}

\usepackage{amsmath,amssymb,amsfonts}
\usepackage{makecell}
\usepackage{subfigure} 
\usepackage{textcomp}
\usepackage{stfloats}
\usepackage{url}
\usepackage{verbatim}
\usepackage{graphicx}
\usepackage{newfloat}
\usepackage{listings}
\usepackage{hyperref}
\usepackage[T1]{fontenc}
\usepackage{aecompl}
\usepackage{booktabs}
\usepackage{multirow} 
\usepackage{color,xcolor}
\usepackage{cleveref}
\usepackage{cite}
\def\BibTeX{{\rm B\kern-.05em{\sc i\kern-.025em b}\kern-.08em
    T\kern-.1667em\lower.7ex\hbox{E}\kern-.125emX}}
\usepackage{balance}
\begin{document}

\title{Hierarchical Reinforcement Learning for Swarm Confrontation with High Uncertainty}

\author{Qizhen Wu, Kexin Liu, Lei Chen, \IEEEmembership{Member, IEEE}, and Jinhu L\"u, \IEEEmembership{Fellow, IEEE}
\thanks{*This work was supported in part by the National Key Research and Development Program of China under Grant 2022YFB3305600, and in part by the National Natural Science Foundation of China under Grants 62141604, 62088101, and 62003015. ({\it{Corresponding
author: Lei Chen.}})}
\thanks{Qizhen Wu, Kexin Liu, and Jinhu L\"u are with the School of Automation Science and Electrical Engineering,
        Beihang University, Beijing 100191, China. (e-mail: wuqzh7@buaa.edu.cn; skxliu@163.com; jhlu@iss.ac.cn)}%
\thanks{Lei Chen is with the Advanced Research Institute of Multidisciplinary Sciences and State Key Laboratory of CNS/ATM, Beijing Institute of Technology, Beijing 100081, China. (e-mail: bit$\_$chen@bit.edu.cn)}%
\thanks{The experiment video is available at \url{https://www.bilibili.com/video/BV15Ts7e8ERZ/?vd_source=9de61aecdd9fb684e546d032ef7fe7bf}.}%
\thanks{The code is available at \url{https://github.com/Wu-duanduan/Swarm_confrontation_HRL}.}%
}

\markboth{Journal of \LaTeX\ Class Files,~Vol.~14, No.~8, August~2021}%
{Shell \MakeLowercase{\textit{et al.}}: A Sample Article Using IEEEtran.cls for IEEE Journals}


\maketitle

\begin{abstract}
    In swarm robotics, confrontation including the pursuit--evasion game is a key scenario. 
    High uncertainty caused by unknown opponents' strategies, dynamic obstacles, and insufficient training complicates the action space into a hybrid decision process.
    Although the deep reinforcement learning method is significant for swarm confrontation since it can handle various sizes, as an end--to--end implementation, it cannot deal with the hybrid process. 
    Here, we propose a novel hierarchical reinforcement learning approach consisting of a target allocation layer, a path planning layer, and the underlying dynamic interaction mechanism between the two layers, which indicates the quantified uncertainty.
    It decouples the hybrid process into discrete allocation and continuous planning layers, with a probabilistic ensemble model to quantify the uncertainty and regulate the interaction frequency adaptively. 
    Furthermore, to overcome the unstable training process introduced by the two layers, we design an integration training method including pre--training and cross--training, which enhances the training efficiency and stability.
    Experiment results in both comparison, ablation, and real--robot studies validate the effectiveness and generalization performance of our proposed approach. 
    In our defined experiments with twenty to forty agents, the win rate of the proposed method reaches around ninety percent, outperforming other traditional methods.
    
\end{abstract}

\def\abstractname{Note to Practitioners}
\begin{abstract}
With artificial intelligence rapidly developing, robots will play a significant role in the future. Especially, the swarm formed by many robots holds promising potential in civil and military applications. Promoting the swarm into games or battles is rather riveting. The reinforcement learning method provides a plausible solution to realize the battle of robotic swarms. There are still some issues that need to be addressed. On one hand, we focus on the uncertainty caused by the battlefield nature and the environment which limits our ability for the implementation of swarms. On the other hand, we solve the problem that the decision process combined with commands and actions is a hybrid system, which cannot be directly reflected in the confrontation of swarms. 
Overall, our approaches throw light on artificial general intelligence and also reveal a solution to interpretable intelligence.

\end{abstract}

\section{Introduction}
\IEEEPARstart{W}{ith} the emergence of artificial intelligence, robotics \cite{guo2024powerful,zhu2023deep} is gaining more attention. Confrontation \cite{xia2023dynamic,piao2023spatio} is a crucial application of robotic swarm, where robots are expected to win through artificial intelligent decision--making. Typical scenarios include pursuit--evasion \cite{li2023optimal} and defense--attack \cite{liu2024game} games. However, its intrinsic mechanism is an $NP$--hard complex problem due to vast involving agents and strong conflict uncertainties \cite{zhang2022improving,xia2021multi,wang2023weighted}. 
Traditional algorithms \cite{hou2023hierarchical,vamvoudakis2022nonequilibrium,xia2023dynamic} are struggling with computation complexities and resource costs as the increasing number of robots and actions.

Deep reinforcement learning (DRL) \cite{kober2013reinforcement} is a plausible solution to this problem. It adopts an end--to--end framework to approach the optimal decision instead of enormous iterations, leading to many accomplishments in racing \cite{kaufmann2023champion} and competition \cite{haarnoja2024learning}.
Driven by maximizing cumulative values in reward functions, DRL optimizes the decision network to produce a desirable strategy. In practice, however, complex problem solving \cite{eichmann2020exploring} brings a new challenge for DRL. 
Focusing on global goals instead of reasoning the problem results in a sparse reward issue \cite{wang2024hybrid,he2023robotic} that limits the application of DRL. 

To avoid constructing intrinsic rewards directly, many researchers employ hierarchical reinforcement learning (HRL)\cite{eppe2022intelligent}, which decouples the complex problem through a divide--and--conquer framework\cite{mao2024dl}. 
The upper layer in HRL divides the timeline into several non--uniform sections each of which it designs a unique reward for the lower layer to train, and, by doing this, the global reward for the upper layer is maximized.  
Sectional rewards fill the timeline leading to the sparse reward problem being solved.
Therefore, designing the unique rewards, also addressed as interaction mechanisms, is a significant trick to HRL.
As a pioneer work, \cite{vezhnevets2017feudal} generates unique rewards from the upper layer in the form of differentiable functions.
It soon shows great potential in dynamic multiple object traveling salesman problem when \cite{guan2022hierrl} investigates the distributed system Ray belonged to {\it{UC Berkeley RISELab}}.
Nevertheless, its framework is totally direct from the upper to the lower. This open--loop feature is inapplicable in many cases, where the performance of the lower layer is not considered once the unique rewards are designed.
For the purpose of the close--loop feature, \cite{ma2021hierarchical} introduces bi--direct layers adjusting the strategy of the upper according to the performance of the lower.
In addition, it limits our ability to promote HRL only by manipulating the lower layer under the command of the upper layer. Hence, \cite{geng2021hierarchical} facilitates an additional reward into the lower to achieve more goals besides unique rewards from the upper. It throws light on a more flexible solution for HRL in other practical problems.

Swarm confrontation, being a typical complex problem solving, naturally consists of discrete and continuous spaces under an uncertain environment. Illustratively, commands on the battlefield always exist in the form of discrete decisions, while actions usually take place in continuous time. 
Recent works \cite{asgharnia2020deception,nian2024large} combine commands and actions into multiple high--level spaces. This method blurs the interpretability of the spaces inevitably resulting in slowly converging algorithms for large--scale swarms. 
Alternatively, it is not hard to reflect the swarm confrontation to divide--and--conquer framework, where the commands are translated into target allocation and the actions are addressed as path planning, from the artificial intelligence perspective\cite{hou2023hierarchical}. Wang et al. \cite{wang2021uav} make an interesting attempt for the first time to introduce HRL into swarm confrontation. Inspired by the method, \cite{kong2023hierarchical2} decomposes multi--aircraft formation air combat into high--level strategy and sub--strategy, achieving favorable effects in confrontation among a few robots. However, the fact that we cannot guarantee the stability of the algorithms prevents us from building a bridge between HRL and swarm confrontation.  
Others \cite{ren2021enabling, mao2024dl} have already improved the stability of the algorithm for HRL in a different scenario. They design an interactive training strategy, including pre--training, intensive training, and alternate training, to ensure the stability guaranteed. 
A prerequisite for this strategy is only the lower layer pre--trained, while the training of the upper is limited by the cumulative feedback from the lower. Thus we cannot directly implement it to swarm confrontation for the environmental uncertainty will make the global optimization impossible.

Here, we propose a guaranteed stable HRL method, which induces quantified uncertainty into an interaction mechanism linking the allocation and planning layers, to solve the hybrid problem of swarm confrontation in various sizes and environments.
Firstly, we construct two--layer DRL networks to reflect commands and actions into target allocation and path planning, respectively, since the high uncertainty caused by the nature of the confrontation, including variant opponents' strategies and transient battlefield environment, demands a hybrid, flexible, and robust intelligent algorithm.
Secondly, the mechanism, which is embedded with a probabilistic ensemble model \cite{janner2019trust} quantifying the uncertainty, regulates the interaction frequency between the two layers. The essence is the frequency is increased as the circumstance becomes uncertain and dire.
Thirdly, a novel integration training method (ITM), consisting of pre--training and cross--training, ensures the stability of HRL, in which, notably, we combine pre--trained and an improved model--based value expansion (IMVE) \cite{buckman2018sample} method together to fasten the convergent speed of the upper in case of few samples given by the lower.
Finally, extensive experiments on different--size swarms verify that our approach outperforms the baselines including non--learning approaches and traditional DRL, and they also demonstrate the necessity of the adaptive frequency approach and ITM through ablation studies. Plus, our method shows that a trained model under a small scale holds favorable generalization in various scales of swarm confrontation.
The main contributions of this paper are summarized as follows.
\begin{itemize}
\item Unveiling the intrinsic mechanism of confrontation, we reflect discrete commands and continuous actions into target allocation and path planning, respectively, and then propose a novel HRL framework including discrete and continuous networks for allocation and planning.

\item We explore that high uncertainty in confrontation is an influential factor between commands and actions, so it is necessary to embed a probabilistic ensemble model by regulating the frequency adaptively for both connecting the two networks and overcoming the uncertainty.

\item Since traditional training methods may be unstable for the HRL framework in our case, we present ITM to ensure the stability of HRL, where we pre--train the commands and actions networks independently and cross--train the two networks facilitating the adaptive interaction mechanism.
\end{itemize}

We organize the rest of the paper as follows. Section \ref{sec:sample2} introduces the research related to our study. Section \ref{sec:sample3} presents the formulation of our problem and provides the preliminaries on DRL. Section \ref{sec:sample4} offers a detailed description and implementation of our two--layer networks and guaranteed stable HRL method. Section \ref{sec:sample5} describes the experiments of our method. Section \ref{sec:sample6} presents our conclusions.

\section{Related Works}
\label{sec:sample2}
As a key scenario of robotics, swarm confrontation is a combinatorial optimization problem with a hybrid decision process and transient environment. This section reviews the related works for swarm confrontation in terms of various methods. We first introduce and analyze the traditionally relevant expert system, game theory, and heuristic approaches, followed by a review of DRL solving swarm confrontation. Moreover, we summarize the characteristics of the existing solutions and further induce the hierarchical learning method for the swarm confrontation problem.

The expert system method \cite{hou2023hierarchical,wang2023decision} models the system with prior knowledge of human experts and selects the strategies in the knowledge base by fuzzy matching approach. 
The method relies on rules developed by an enormous number of human experts and is unable to ensure the optimality of decisions in a complex confrontation environment. In game theory\cite{vamvoudakis2022nonequilibrium,liu2022distributed}, the swarm confrontation problem is frequently modeled as a differential game. 
It suffers from problems such as too many state variables and complex differential equations, making it difficult to apply to complex multi--agent environments. The heuristic approach \cite{xia2023dynamic,liu2022evolutionary} considers modeling the swarm as biologically inspired networks to simulate the dynamics of swarm confrontation, which has more potential to solve large--scale confrontational problems.
However, the simulation and test require a lot of computing resources and time.

Without relying on prior knowledge, DRL learns strategies by interacting with the environment. As a result, it gains more attention in fields such as game playing\cite{wurman2022challenges}, natural language processing\cite{gao2023scaling}, and robotics\cite{kober2013reinforcement}. Recently, DRL has been applied to swarm confrontation, where it learns rules from huge numbers of problem instances rather than designing them manually. De et al.\cite{de2021decentralized} combine DRL with curriculum learning for pursuing an omnidirectional target with multi--agent. However, taking the single--agent DRL approach becomes difficult when faced with large--scale swarm confrontation scenarios. Therefore, Xia et al.\cite{xia2021multi} propose an end--to--end multi--agent reinforcement learning to enable agents to make decisions for cooperative target tracking. Qu et al.\cite{qu2023pursuit} further provide an adversarial--evolutionary game training method and designed obstacle avoidance scenarios in swarm confrontation.

Among the existing methods for solving the swarm confrontation problem, expert system, game theory, and heuristic methods cannot be well applied to large--scale problems. The performance of all three methods degrades significantly when the problem size increases. DRL is a desirable alternative due to the ability to learn strategies just by interacting with the environment. However, the direct use of an end--to--end approach in swarm confrontation \cite{asgharnia2020deception,nian2024large}, results in the non--interpretability of the hybrid decision space, hindering the training of DRL on large--scale swarms.

To deal with the above issues, the hierarchical learning method reflects the swarm confrontation to divide-–and–-conquer framework, where the commands and the actions are addressed as discrete and continue decision spaces, respectively. Hou et al.\cite{hou2023hierarchical} integrate the finite state machine and event--condition--action frameworks to give a more interpretable solution. Wang et al.\cite{wang2021uav} decompose the swarm confrontation problem into multiple tasks to reduce the challenges of sparse reward learning. Kong et al.\cite{kong2023hierarchical1} employ the goal--conditioned HRL framework with feedback and construct a dual--aircraft formation air combat scenario. They make interesting attempts to introduce HRL into swarm confrontation. Inspired by them, we design a guaranteed stable HRL method for quantifying the uncertainty caused by unknown opponents' strategies, moving obstacles, and insufficient training. It establishes a dynamic interaction mechanism between the upper and lower layers, which has favorable potential for solving the realization problem of swarm confrontation.

\section{Problem Formulation and Preliminaries}
\label{sec:sample3}
\subsection{Definition of Swarm Confrontation}
\label{sec:sample3A}
\begin{figure}[t]
    \centering
    \includegraphics[width=0.48\textwidth]{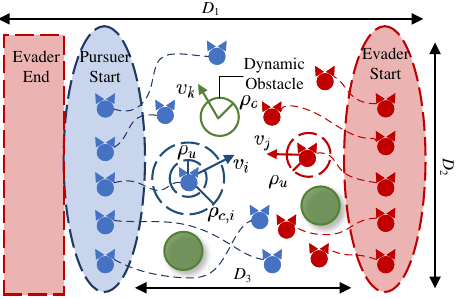}
    \caption{Representation of the parameters involved in the pursuit--evasion game.}
    \label{fig1}
\end{figure}

This study considers the swarm confrontation as a pursuit--evasion game where exists dynamic obstacles. There are multiple pursuers under different abilities' constraints to jointly guard the preset target point. They need to work together to capture evaders as soon as possible while avoiding collisions with obstacles and neighbors. Evaders need to get to the preset target point as quickly as possible without being caught by pursuers, as shown in Fig.\ref{fig1}. We use blue and red agents to denote pursuers and evaders, respectively. $D_1$, $D_2$, and $D_3$ denote the length and width of the battlefield area and the straight--line distance between the starting areas of both sides, respectively. To simplify the movement of agents, we consider the pursuit--evasion game in a two--dimensional scenario. Let position and velocity vectors in two dimensions be denoted by $p$ and $v$, respectively. Pursuers and evaders cannot exceed the boundaries of the scenario and the maximum dynamic constraints, which can be bounded as
\begin{align}
    -\frac{D_{1}}{2}\leq p_x\leq \frac{D_{1}}{2},  \nonumber \\
    -\frac{D_{2}}{2}\leq p_y\leq \frac{D_{2}}{2}, \nonumber \\ 
    \lVert v \rVert \leq \lVert v \rVert _{\max}
    \label{equ0},
\end{align} 
where $\lVert \cdot \rVert$ denotes the Euclidean norm. $p_x$ and $p_y$ denote the positions of the horizontal and vertical coordinates, respectively. $\lVert v \rVert _{\max}$ denotes the maximum velocity magnitude.
Let $U_p=\left\{ u_{p,i}|i=1,...,I \right\} $ and $U_e=\left\{ u_{e,j}|j=1,...,J \right\}$ denote a swarm of pursuers and evaders, respectively. Both pursuers and evaders should avoid dynamic obstacles $O=\left\{ o_{k}|k=1,...,K \right\}$. For any pursuer $u_{p,i}=\left( p, v, \rho_c, \lVert v \rVert _{\max} \right)$, $\rho_c$ denotes the capture radius. For any evader $u_{e,j}=\left( p, v, \lVert v \rVert _{\max} \right)$, its maximum velocity magnitude is greater than the pursuers. We consider the agent and obstacle as the circle models with radius $\rho_u$ and $\rho_o$, respectively. In this paper, we design the HRL method for pursuers, while evaders use the artificial potential field method \cite{de2021decentralized}. The HRL method reflects the commands and actions of the pursuit--evasion game into target allocation and path planning, respectively, through the divide--and--conquer framework approach. We can use $x_{ij}$ to denote the allocation variable which represents whether to allocate $u_{p,i}$ to $u_{e,j}$. It yields that
\begin{equation}
    \begin{array}{c}
        x_{ij}=\left\{ \begin{array}{c}
	1,\ \text{if\ }u_{p,i}\ \text{is\ allocated\ to\ }u_{e,j}\\
	0,\ \text{otherwise}\\
\end{array} \right. .
    \end{array}\label{equ1}
\end{equation}
$\left[ x_{ij} \right] ,\ i\in I,\ j\in J$ denotes the allocation matrix for all pursuers. Subsequently, pursuers chase the assigned target through real--time path planning. Throughout the process, target allocation and path planning are dynamically alternated. For pursuers and evaders, we design the rules as follows: 
\begin{itemize}
\item[1)]Each pursuer is allowed to select only one evader, and one evader can be assigned to several pursuers.
\item[2)]If an evader enters the capture radius of one of the pursuers, it is being captured.
\item[3)]Both evaders and pursuers that satisfy rule 2) cannot continue moving.
\item[4)]The successful condition of pursuers: evaders do not reach the preset target point within the specified time or more than half of evaders are captured.
\item[5)]The successful condition of evaders: more than half of evaders can reach the predetermined target point within the time limit without being captured by pursuers.
\end{itemize}

Rule 3) explains that a pursuer can capture up to one evader. We set up that there is the same number of pursuers and evaders in the pursuit--evasion game. If pursuers are less than evaders, pursuers cannot capture all the evaders in this setting. Otherwise, if pursuers are more than evaders, it will be not critical for pursuers to develop a strategy to capture as many evaders as possible.

\subsection{Markov Decision Process}
Let $\mathbb{R}$ denote the set of real numbers. $\mathbb{E}$ denotes the mathematical expectation.
We can model reinforcement learning by a Markov decision process (MDP). We represent the MDP as a five--tuple $\left( S,A,P,R,\gamma \right)$. $S$ and $A$ denote the state of the environment and action of the agent, respectively. $P\left( s,a  \right) :S\times A\times S\rightarrow [0,1]$ denotes the state transition probabilities, $R:S\times A\rightarrow \mathbb{R}$ denotes the reward function, and $\gamma$ denotes the discount factor. 

MDP is the case when there is only a single agent or when the system is considered a centralized agent. We can describe a fully cooperative multi--agent reinforcement learning task as a decentralized partially observable Markov decision process (Dec--POMDP). We represent Dec--POMDP as a six--tuple $\left(S,A,P,R,Z,\gamma \right)$, where the state space $S$, state transition function $P$, reward function $R$ and the discount factor $\gamma$ have the same denotation as the MDP. For each agent $i$, $a^i\in A$ and $a = \left[ a^1,\cdots ,a^I \right]$ denote the action of each agent and the set of the joint actions of all agents, respectively. $z^i\in Z$ and $z = \left[ z^1,\cdots ,z^I \right]$ denote the observation of each agent and the set of the joint observations of all agents, respectively. $\varGamma^i =\left( z_0^i,a_0^i,z_1^i,a_1^i,... \right)$ is the trajectory of each agent interacting with the environment under the strategy, where $a_t^i\sim\pi^i \left( z_t^i \right)$ and $z_{t+1}^i\sim P\left( z_t^i,a_t^i \right)$ denote the selected action and reaching observation at each decision step $t$, respectively. The purpose of each agent is to optimize the policy network $\pi^i$ such that the cumulative rewards $\mathbb{E}_{\varGamma^i \sim\pi^i }\left[ \sum_{t=1}^{\infty}{\gamma ^tr_{t}} \right] ,r_{t} \sim R$ is maximized under the policy.

\section{Methodology}
\label{sec:sample4}
In this section, we introduce the guaranteed stable HRL method for solving the swarm confrontation problem in the dynamic obstacles environment. In the upper layer, it constructs an MDP model and designs a centralized deep Q--learning (DQN) algorithm for the target allocation. In the lower layer, it establishes a Dec--POMDP model and proposes a multi--agent deep deterministic policy gradient (MADDPG) algorithm for path planning. Then, the method feeds the cumulative rewards from the lower to the upper, and adopts a probabilistic ensemble network to quantify the uncertainty caused by unknown evaders' strategies, moving obstacles, and insufficient training. Based on the uncertainty quantification, we use an adaptive truncation method to optimize the interaction frequency between the two layers. In addition, we employ an improved model–based value expansion method to enhance the sample utilization in the upper layer which has fewer samples. Afterward, we design an integration training method including pre--training and cross--training to enhance the training efficiency and stability of our approach.

\subsection{Upper Layer for Target Allocation}
\begin{itemize}
\item[1)]\textbf{Markov Decision Process}. The procedure of the target allocation in the upper layer can be deemed as a sequential decision--making process, where a pursuer will be assigned for each evader. We cast such a process as an MDP which includes state space $S$, action space $A$, reward $R$, and state transition $P$. The detailed definition of our MDP is stated as follows.

\item[$\bullet$]\textbf{State}. The state mainly consists of the current allocation of all pursuers $\left[ x_{ij} \right]$ as well as information on pursuer $i$ that currently needs to be allocated. It can be given as
\begin{align}
        s = \left( \left[ x_{ij} \right],p_{i},v_{i},\rho_{c,i}, \lVert v \rVert_{\max,i} \right) .
    \label{equ2}
\end{align}

\item[$\bullet$]\textbf{Action}. The action consists of information on the selected evader, which is
\begin{align}
        a =  \left( p_{j},v_j,\lVert v \rVert_{\max,j} \right)  .
    \label{equ3}
\end{align}

\item[$\bullet$]\textbf{Reward}. 
The capture probability of pursuer $i$ relative to evader $j$ can be assessed by the distance between them and the capture radius, which is described as
\begin{align}
        q_{ij} =  \frac{\rho _{c,i}}{\rho _{c,i}+\lVert p_i-p_j \rVert} .
    \label{equ41}
\end{align}
If there are multiple pursuers assigned to the same evader $j$, then the joint capture probability is
\begin{align}
        \overline{q}_j=1-\prod_i{\left( 1-q_{ij} \right) , ~ i}\in \left\{ k\left| x_{kj}=1 \right. \right\} .
    \label{equ42}
\end{align}
The capture value of the evader is related to its maximum velocity magnitude, therefore, the goal of target allocation can be set to maximize the following effectiveness function:
\begin{align}
    \mathbb{E}\left( \left[ x_{ij} \right] \right) &=\sum_1^J{\overline{q}_j \lVert v \rVert_{\max,j}} \nonumber \\
&=\sum_1^J{\left[ 1-\prod_i{\left( 1-q_{ij} \right)} \right] \lVert v \rVert_{\max,j}}.
\label{equ43}
\end{align}
Let $\left[ \tilde{x}_{ij} \right]$ denote the allocation matrix after allocating pursuer $i$ to evader $j$. The local reward in allocation can be obtained as
\begin{align}
        r_{\text{allo},1} = \mathbb{E}\left( \left[ \tilde{x}_{ij} \right] \right) - \mathbb{E}\left( \left[ x_{ij} \right] \right) .
    \label{equ44}
\end{align}
Once all the pursuers have completed their assignments, we get the global reward in allocation, which can be calculated as
\begin{align}
        r_{\text{allo},2} = \frac{1}{I}\mathbb{E}\left( \left[ x_{f} \right] \right)  ,
    \label{equ45}
\end{align}
where $\left[x_{f} \right]$ denotes the allocation matrix in final.
The static reward for target allocation is a linear operation consisting of the local and global allocation rewards, which can be expressed as follows:
\begin{align}
r_{\text{allo}}= \omega_1 r_{\text{allo},1} + \left(1 - \omega_1 \right) r_{\text{allo},2},
    \label{equ4}
\end{align}
where $\omega_1$ is the weighting factor between the local and global allocation rewards.
\item[$\bullet$]\textbf{State transition}. After pursuer $i$ selects evader $j$, $x_{ij}$ becomes one, and it is the turn of pursuer $i+1$ for target allocation.

\item[2)]\textbf{Training method}. Combining the above MDP settings, we adopt double DQN for training on target allocation. The method estimates the optimal state--action value function $Q^*:S\times A\rightarrow \mathbb{R}$ through a parameterized neural network $Q_{\theta}\left( s_t,a_t \right) \approx Q^*\left( s_t,a_t \right) =\mathbb{E}\left[ R\left( s_t,a_t \right) +\gamma \max Q^*\left( s_{t+1},a_{t+1} \right) \right] ,\forall s_t\in S$. The subscript $\theta$ is the weighting factor of the value--function network. For $\gamma \approx 1$, $Q^*$ estimates the discounted returns of the optimal strategy over an infinite range. The method approximates $Q^*$ by $Q_{\theta}$ and the loss function is set in the following form:
\begin{equation}
    \begin{array}{c}
        L_{\theta}=\mathbb{E}_{\left( s_t,a_t \right) \sim \mathcal{D}}\lVert Q_{\theta}\left( s_t,a_t \right) -y \rVert ^2,
    \end{array}\label{equ5}
\end{equation}
where $y=R\left( s_t,a_t \right) +\gamma \max Q_{\theta ^-}\left( s_{t+1},a_{t+1} \right)$ is the $Q$--target. The subscript $\theta^{-}$ is a slow--moving online average that avoids overestimation of $Q^*$. At each iteration, it is updated with the following rule:
\begin{equation}
    \begin{array}{c}
        \theta _{t+1}^{-}\gets \left( 1-\zeta \right) \theta _{t}^{-}+\zeta \theta _t,
    \end{array}\label{equ6}
\end{equation} 
where $\zeta \in \left[ 0,1 \right)$ is a constant factor. $\mathcal{D}$ is a replay buffer that iteratively grows as data are updated. 
We use the same DQN network with a forward inference structure as \cite{luo2021learning}, thus decoupling the network from the size of the problem, which is similar to the critic network in deep deterministic policy gradient (DDPG). Based on centralized DQN, the pursuers maximize the cumulative rewards in (\ref{equ4}) with a cooperative approach for optimal target allocation.
\end{itemize}

\subsection{Lower Layer for Path Planning}
\begin{itemize}
\item[1)]\textbf{Decentralized Partially Observable Markov Decision Process}. Based on the allocation result of the upper layer, each pursuer needs to plan a route to chase the assigned evader and avoid collisions. We model this process as a Dec--POMDP, including the state space $S$, observation space $Z$, action space $A$, reward $R$, and state transition $P$:
\item[$\bullet$]\textbf{State}. The global state space includes information on all pursuers and evaders. It can be described as
\begin{align}
    s=\left(U_P, U_e\right),
        \label{equ6_1}
\end{align}

\item[$\bullet$]\textbf{Observation}. Considering the fast--moving characteristics of agents, instead of setting fixed distance thresholds to construct an observation vector, it is assumed that agents can observe the nearest neighbor. The local observation of a pursuer contains information on the allocated evader, its nearest neighbor, and obstacle. It yields that
\begin{align}
    z=\left(p_i,p_j,p_i^n,p_k,v_i,v_j,v_i^n,v_k\right),
        \label{equ7}
\end{align}
where $p_i^n$ and $v_i^n$ are the position and velocity of the nearest neighbor of pursuer $i$.

\item[$\bullet$]\textbf{Action}. The action consists of the velocity magnitude $\lVert v \rVert \in [0,\lVert v \rVert _{\max}]$ and velocity direction $\psi \in [-\pi, \pi]$ of pursuer $i$, which is denoted as
\begin{align}
        a=\left( \lVert v_i \rVert, \psi_i \right).	
    \label{equ8}
\end{align}

\item[$\bullet$]\textbf{Reward}. 
To intercept the allocated evader, pursuer $i$ receives an intrinsic reward from the upper layer during path planning. The intrinsic reward can be described as
\begin{align}
r_{\text{int}}=\left\{ \begin{matrix}
	\displaystyle -\frac{\lVert p_i-p_j \rVert}{\rho_{c,i}}+r_a&		\lVert p_i-p_j \rVert<\rho_{c,i}\\
	\displaystyle -\frac{\lVert p_i-p_j \rVert}{\rho_{c,i}}&		\text{otherwise}\\
\end{matrix} \right..
    \label{equ9}
\end{align}
In path planning, pursuers need to avoid collisions with their neighbors and obstacles while chasing their allocated targets. The avoidance reward is set in the following form:
\begin{align}
        r_{\text{avo}}=r_{\text{avo},1}+r_{\text{avo},2}.
    \label{equ10}
\end{align}
If pursuer $i$ enters the threat zone of its nearest obstacle $k$ ($\lVert p_i-p_k \rVert<\rho_u + \rho_o+d_{\text{thr}}$), there is
\begin{align}
r_{\text{avo},1}=\frac{\lVert p_i-p_k \rVert-\left( \rho _u+\rho _o+d_{\text{thr}} \right)}{\left( \rho _u+\rho _o+d_{\text{thr}} \right)}-r_b,
    \label{equ11}
\end{align}
otherwise, $r_{\text{avo},1}=0$. And if pursuer $i$ enters the threat zone of its nearest neighbor ($2\rho _u<\lVert p_i-p_i^n \rVert<2\rho _u+d_{\text{thr}}$), there is
\begin{align}
r_{\text{avo},2}=\frac{\lVert p_i-p_i^n \rVert-\left( 2\rho _u+d_{\text{thr}} \right)}{\left( 2\rho _u+d_{\text{thr}} \right)}-r_c,
    \label{equ12}
\end{align}
otherwise, $r_{\text{avo},2}=0$.
$r_a,r_b,r_c$ are constant rewards. We set a threat distance $d_{\text{thr}}$ to avoid collisions between pursuer $i$ with its neighbors and obstacles. The final reward for path planning is a linear operation consisting of the intrinsic reward and avoidance reward, which can be calculated as
\begin{align}
        r_{\text{path}}=\omega_2 r_{\text{int}}+ \left(1 - \omega_2 \right) r_{\text{avo}}.
    \label{equ13}
\end{align}
where $ \omega_2$ is the weighting factor between the intrinsic reward and avoidance reward. 
\item[$\bullet$]\textbf{State transition}. Based on the current state and decisions, we can calculate the position at the next moment through the first--order dynamics model
\begin{align}
        p_{x,t+1} = p_{x,t} + \delta t \lVert v \rVert  \cos \psi, \nonumber \\
        p_{y,t+1} = p_{y,t} + \delta t \lVert v \rVert  \sin \psi,
    \label{equ13_1}
\end{align}
where $\delta t$ is the length of time step.
\item[2)]\textbf{Training method}.
To enable pursuers to make decisions based on local observations while realizing collaboration with each other to accomplish interception, we use the MADDPG algorithm, which is a centralized training with decentralized execution method. It designs a separate critic network $Q_{\theta}^{i}\left( z_t,a_t \right)$ for each agent, which is updated similarly to (\ref{equ5}) for double DQN as follows:
\begin{equation}
    \begin{array}{c}
        L_{\theta}^{i}=\mathbb{E}_{\left( z_t,a_t \right) \sim \mathcal{D}}\lVert Q_{\theta}^{i}\left( z_t,a_t \right) -y^{i} \rVert ^2,
    \end{array}\label{equ14}
\end{equation}
where $y^{i}=R^{i}\left( z_t,a_t \right) +\gamma \max Q_{\theta ^-}^{i}\left( z_{t+1},a_{t+1} \right)$. In addition, each agent holds a policy network $\pi_{\beta}^{i}\left( z^i_t\right)$ which has the following loss function:
\begin{equation}
    \begin{array}{c}
        L_{\beta}^{i}=-\mathbb{E}_{\left( z_t,a_t \right) \sim \mathcal{D}}\left[Q_{\theta}^{i}\left( z_t,a_t \right)\right].
    \end{array}\label{equ15}
\end{equation}
The subscript $\beta$ is the weighting factor of the policy network. Based on MADDPG, the pursuers maximize the cumulative rewards in (\ref{equ13}) by collaborating while autonomous path planning.

\end{itemize}

\begin{figure*}[t]
    \centering
    \includegraphics[width=0.8\textwidth]{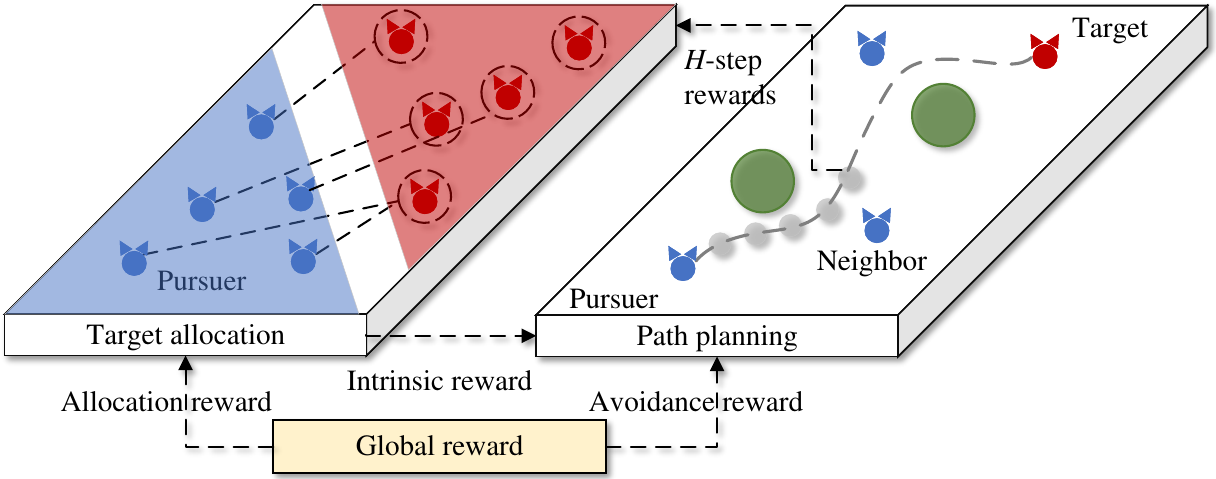}
    \caption{Structure of the two--layer networks under swarm confrontation.}
    \label{fig2}
\end{figure*}

\subsection{Hierarchical Network Interaction Method}
This study decouples target allocation and path planning into a two--layer networks, and unifies them into a dynamic cyclic process. After $H$ time steps in path planning, we need to redo the target allocation based on the current state, where interaction step $H$ is a variable related to the current state $s_t$ and allocation $a_t$. In this dynamic process, the upper layer allocates targets and provides an intrinsic reward to the lower layer, which chases the assigned targets while avoiding obstacles through real--time path planning. 
To quantify the uncertainty including variant opponents’ strategies and transient battlefield environment, we construct a virtual environment model $\mathcal{M}_\phi$ that incorporates both state transition and reward function, which is expressed as
\begin{align}
        \hat{s}_{t+H},\hat{r}_{t} \gets \mathcal{M}_\phi \left(s_t , a_t \right),
    \label{equ16}
\end{align}
where $\hat{s}_{t+H},\hat{r}_{t}$ denote the predicted value of $s_{t+H}$ and $r_{t}$ with the environment model, respectively. 
We use an ensemble neural network $( m_\phi^1,...,m_\phi^B )$ proposed in \cite{janner2019trust}, which can be used to quantify the epistemic and aleatoric uncertainty in the environment. Epistemic uncertainty
results from the lack of sufficient training in the lower layer and aleatoric uncertainty refers to the unknown enemies' strategies and moving obstacles in this study. It takes the state--action pair as input and outputs Gaussian distribution $\mathcal{N}$ of the next state and reward. The model $m_\phi^b$ can be expressed in the following form:
\begin{align}
        m_\phi^b\left( s_{t+H},r_t\mid s_t,a_t \right) =\mathcal{N}\left( \mu _{\phi}^b\left( s_t,a_t \right) ,\sigma _{\phi}^b\left( s_t,a_t \right) \right),
    \label{equ17}
\end{align}
where $\mu$ and $\sigma$ represent the mean and variance of the Gaussian distribution, respectively. This transition dynamics model is trained to maximize the expected likelihood
\begin{align}
    \displaystyle L_{\phi}^b=\mathbb{E}_{\left(s_{t}, a_{t}\right) \sim \mathcal{D}}\frac{1}{\lVert \mathcal{D} \rVert} \left[\frac{\log \sigma_{m_{\psi}^{b}}}{2}
    +\frac{\left(d_{t}-\mu_{m_{\psi}^{b}} \right)^{2}}{2 \sigma_{m_{\psi}^{b}}} \right],
    \label{equ18}
\end{align}
where $d_t$ represents the model outputs. We omit the inputs $\left(s_{t}, a_{t}\right)$ of $\mu_{m_{\psi}^{b}}$ and $\sigma_{m_{\psi}^{b}}$ for brevity. Then, we employ the outputs of sub--models to denote the mean and variance of the ensemble model $\mathcal{M}_\phi$, which are calculated as
\begin{align}
    \mu_{\mathcal{M}_\phi}= & \frac{1}{B} \sum_{b=1}^{B}\mu_{m_{\psi}^{b}}, \nonumber \\
    \sigma_{\mathcal{M}_\phi}^{2}= & \frac{1}{B} \left[\sum_{b=1}^{B} \left( \sigma_{m_{\psi}^{b}}^{2} + \mu_{m_{\psi}^{b}}^{2} \right) -\mu_{\mathcal{M}_\phi}^{2}\right].
    \label{equ19}
\end{align} 
The uncertainty of $\mathcal{M}_\phi$ can be set to the variance of the ensemble model. It is denoted as
\begin{align}
    \hat{V}\left(s_{t}, a_{t}\right) = \sigma_{\mathcal{M}_\phi}^{2}\left(s_{t}, a_{t}\right).
    \label{equ20}
\end{align}

Instead of the conventional fixed--step or infinity iteration method, we adopt an adaptive truncation approach, which calculates the prospective value of $H$ by the following linear operation:
\begin{align}
    H\left(s_t, a_t\right)= \lfloor -\omega_3 \hat{V}\left(s_t, a_t \right)+H_{\text{base}} \rfloor,
\label{equ21}
\end{align}
where $\omega_3$ is a weighting factor and $H\left(s_t, a_t\right)$ is an integer limited in $\left[ H_{\min},H_{\max} \right]$. Subscripts $\text{base}$, $\min$, and $\max$ are the settled based, minimum, and maximum values, respectively. $\lfloor x \rfloor =\max \left\{ m\in \mathbb{Z}|m\leqslant x \right\}$, where $x \in \mathbb{R}$ and $\mathbb{Z}$ denotes the set of integers. In \cite{chua2018deep}, it is mentioned that aleatoric uncertainty cannot be reduced with training, but we quantify the uncertainty to adjust the interaction frequency adaptively. When uncertainty $\hat{V}\left(s_t, a_t \right)$ is higher, $H$ is smaller, indicating the need for more frequent target allocation by the upper layer. As the model $\mathcal{M}_\phi$ fits the environment, the epistemic uncertainty in the lower layer decreases, as well as the frequency of target allocation.

In addition, we feed the cumulative rewards to the upper after the lower executes $H$ time steps, as shown in Fig. \ref{fig2}. The static allocation reward in (\ref{equ4}) and the cumulative path rewards in (\ref{equ13}) encourage agents to allocate reasonable targets and plan feasible paths, respectively. Therefore, the upper layer linearly weights the above rewards as the total reward for target allocation, which can be denoted as
\begin{align}
        r_{\text{total}}= H \cdot r_{\text{allo}} + \sum_1^H{r_{\text{path}}} + r_{\text{capt}},
    \label{equ22}
\end{align}
where $r_{\text{capt}}$ is an additional reward equal to the number of captured evaders during these $H$ time steps. Based on the total reward, the upper layer is updated by making a trade--off among the above three objective rewards simultaneously.
However, samples are greatly reduced since the effect of a single allocation in the upper layer can only be obtained from the feedback of the lower layer planning after $H$ steps. Here, we employ an improved model--based value expansion (IMVE) method to improve the sample utilization of the upper layer. The method performs $N$--step value estimation based on the real environment state and allows $Q_\theta$ of $N$ steps to converge to $Q^*$. Similar to $H\left(s_t, a_t\right)$, $N\left(s_t, a_t\right)$ is affect by the uncertainty quantified by the model $\mathcal{M}_\phi$. We can calculate the prospective length of $N$ in the following form:
\begin{align}
    N\left(s_t, a_t\right)= \lfloor -\omega_4 \hat{V}\left(s_t, a_t \right)+N_{\text{base}} \rfloor,
\label{equ23}
\end{align}
where $\omega_4$ is a weighting factor and $N\left(s_t, a_t\right)$ is an integer limited in $\left[ 1,N_{\max} \right]$. The goal of policy in double DQN is to 
\begin{figure*}[t]
    \centering
    \includegraphics[width=0.9\textwidth]{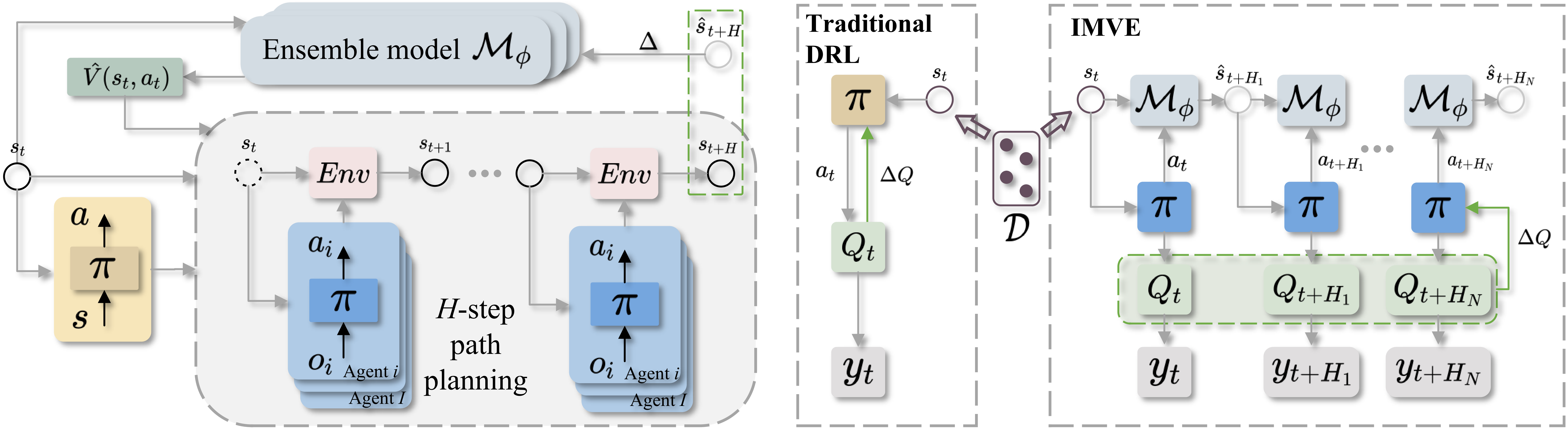}
\caption{Framework of our method and IMVE.}
\label{fig3}
\end{figure*}
maximize the long--term cumulative rewards for all future moments, which can be expressed as
\begin{align}
        \pi_{1}\left( s_t \right) =~\underset{a_{t}}{\arg\max} ~\mathbb{E} \left[ \sum_{m=t}^{\infty}{\gamma ^mR\left( s_m,a_m \right)} \right],
        \label{equ24}
\end{align}
where $\underset{x}{\arg\max}~F(x)$ represents the value of variable $x$ when $F(x)$ obtains its maximum value. We reduce the prediction interval from infinity to a visual step $N$, and replace the reward $R\left( s_m,a_m \right)$ with the cumulative rewards $\sum_{g=m}^{\infty}{\gamma ^gr_g}$ in (\ref{equ24}). 
The strategy of IMVE can be described in the following form:
\begin{equation}
    \begin{array}{c}
        \pi _{2}\left( s_t \right) =\underset{a_{t:t+N}}{\arg\max} ~\mathbb{E}\left[ \sum\limits_{m=t}^{t+N}{\gamma ^m\left( \sum\limits_{g=m}^{\infty}{\gamma ^gr_g} \right)} \right].
    \end{array}\label{equ25}
\end{equation}
It produces a locally optimal solution by performing predictive control based on the single moment $t$. Through rolling iterations at different times, it generates multiple locally optimal solutions and calculates the optimal value. 
The use of cumulative rewards compensates for the lack of the terminal reward. IMVE improves the sample utilization and training efficiency by performing $N$--step value estimation and convergence, the loss function in (\ref{equ5}) is improved as follows:
\begin{equation}
    \begin{array}{c}
        L_{\theta}=\mathbb{E}_{\left( s_t,a_t \right) \sim \mathcal{D}}\sum\limits_{n=0}^{N-1}{\hat{\omega}\lVert Q_{\theta}\left( \hat{s}_{t+H_n},\hat{a}_{t+H_n} \right) -y_{t+H_n} \rVert ^2},
    \end{array}\label{equ26}
\end{equation}
where $H_n$ is defined in the following recursive form:
\begin{align}
        H_n = H \left( s_{t+H_{n-1}}, a_{t+H_{n-1}} \right) + H_{n-1},
        \label{equ12}
\end{align} 
where $H_0=0$. The $Q$--target is modified as follows:
\begin{align}
        \displaystyle y_{t+H_n} = ~& \sum\limits_{i=n}^{N-1}\gamma ^{i-n}~\hat{r}_{t+i} \nonumber\\ 
        \displaystyle & +  \gamma ^{N-n} \max Q_{\theta ^-}\left( s_{t+H_N},a_{t+H_N} \right).
        \label{equ27}
\end{align} 
In (\ref{equ26}), it introduces a discount weight $\hat{\omega}$, which is related to the uncertainty of the model ${M}_\phi$ with respect to the sample $\left( \hat{s}, \hat{a} \right)$ produced. $\hat{\omega}$ is assigned as follows:
\begin{align}
    \hat{\omega}\left(\hat{s}_t, \hat{a}_t \right)= -\omega_5 \hat{V}\left(\hat{s}_t, \hat{a}_t \right)+\hat{\omega}_{\text{base}},
\label{equ28}
\end{align}
where $\omega_4$ is a weighting factor and $\hat{\omega}\left(\hat{s}_t, \hat{a}_t \right)$ is limited in $\left[ \hat{\omega}_{\min},1 \right]$.
By constructing the environment model $M_\phi$, we can quantify the epistemic and aleatoric uncertainty, thus allowing $H$, $N$, and $\hat{\omega}$ to adjust adaptively. Moreover, it allows us to improve the sample utilization of the upper layer based on IMVE. The framework of our method is shown in Fig. \ref{fig3}.

\subsection{Integration Training Method}
Within the HRL, we design the upper layer and lower layer for target allocation and path planning, respectively. The upper layer assigns targets and intrinsic rewards to the lower layer, and the cumulative rewards of the lower are fed to the upper after $H$ time steps. To deal with the strong correlation between the two layers, we propose the integration training method (ITM) consisting of pre--training and cross--training. The training framework is shown in Algorithm \ref{alg1}.

\begin{algorithm}[t]
    \renewcommand{\algorithmicrequire}{\textbf{Input:}}
    \renewcommand{\algorithmicensure}{\textbf{Initialization:}}
	\caption{Integration Training Method}
    \label{alg1}
    \begin{algorithmic}[1] 
        \REQUIRE  the number of upper layer pre-training episodes $E_{p,u}$; the number of lower layer pre-training episodes $E_{p,l}$; the number of upper layer training steps $T_u$; the number of lower layer training steps $T_l$; the number of cross-training episodes $E_c$;
	    \ENSURE  the training datasets of the pre-training and cross-training processes; the parameters of networks in the upper layer $Q_{\theta_u}$, lower layer $Q^i_{\theta_l},\pi^i_{\beta_l}$, and Model $\mathcal{M}_\phi$; 
        
        \FORALL {$e\ =\ 1\rightarrow E_{p,u}$}
            \FORALL {$t\ =\ 1\rightarrow T_u$}
                \STATE $r_u \gets$ Eq. (\ref{equ4});
                \COMMENT{static allocation reward}
                \STATE $L_u \gets$ Eq. (\ref{equ5});
                \STATE Refresh network $Q_{\theta_u}$;
                \COMMENT{gradient-descent}
            \ENDFOR
        \ENDFOR
        \FORALL {$e\ =\ 1\rightarrow E_{p,l}$}
            \FORALL {$t\ =\ 1\rightarrow T_l$}
                \STATE $r_l \gets$ Eq. (\ref{equ13});
                \COMMENT{path planning reward}
                \STATE $L_l \gets$ Eq. (\ref{equ14})--(\ref{equ15});
                \STATE Refresh networks $Q^i_{\theta_l},\pi^i_{\beta_l}$;
            \ENDFOR
        \ENDFOR
        \FORALL {$e\ =\ 1\rightarrow E_c$}
            \STATE $t_0 \gets 1$, $H \gets$ Eq. (\ref{equ21})
            \COMMENT{initialize $H$}
            \FORALL {$t\ =\ 1\rightarrow T_l$}
                \STATE $r_l \gets$ Eq. (\ref{equ13});
                \STATE $L_l \gets$ Eq. (\ref{equ14})--(\ref{equ15});
                \STATE Refresh networks $Q^i_{\theta_l},\pi^i_{\beta_l}$;
                \IF{$t = t_0 + H$}
                    \STATE $r_u \gets$ Eq. (\ref{equ22});
                    \COMMENT{total reward}
                    \STATE $L_u \gets$ Eq. (\ref{equ5});
                    \STATE Refresh networks $Q_{\theta_u}$ and $\mathcal{M}_\phi$;
                    \STATE $t_0 \gets t$, $H \gets$ Eq. (\ref{equ21})
                    \COMMENT{refresh $H$}
                \ENDIF
            \ENDFOR
        \ENDFOR
    \end{algorithmic}
\end{algorithm}

At the beginning of the ITM, we construct the pre--training phase for the upper layer and lower layer for $E_{p,u}$ and $E_{p,l}$ episodes, respectively. This phase allows the upper layer to learn a static allocation strategy and the lower layer to train an initial path planning policy, which supports the subsequent training phase. After the pre--training phase, we cross--train the two--layer networks for $E_c$ episodes. Since the value of $H$ keeps changing according to the state $s$ and allocation result $a$, we have to keep it refreshed. Also, the model $M_\phi$ is keeping updating at the same time in this phase.

Based on this method, we can effectively utilize a large amount of data to learn the initial network in pre--training, thus speeding up the learning process. In addition, it can avoid training instability due to the interactions between the two layers in cross--training.

\section{Experiments}
\label{sec:sample5}
In this section, extensive experiments are conducted to evaluate the performance of our HRL method for solving the swarm confrontation problem in different sizes. In comparison experiments, we adopt various baselines including the expert system, game theory, heuristic, and traditional DRL algorithms. We verify the influence of the adaptive frequency approach and ITM in ablation studies. Moreover, we apply the model trained under a small size to solve larger ones to investigate the generalization of our method. Finally, we deploy the model after integration training in simulations to the real--robot system.

\begin{table}[t]
\renewcommand{\arraystretch}{1.25}
\setlength{\tabcolsep}{16pt}
\centering
\caption{The Hyperparameters of Upper Layer and Lower Layer.}
\label{table1}
\begin{tabular}{|c|c|c|c|}
\hline
\multicolumn{2}{c|}{Parameter} &\multicolumn{1}{c|}{Upper layer} &\multicolumn{1}{c}{Lower layer}\\
\hline

\multicolumn{2}{c|}{Learning rate $\alpha$} & \multicolumn{1}{c|}{$10^{-4}$} & \multicolumn{1}{c}{$10^{-3}$}\\

\multicolumn{2}{c|}{Discount factor $\gamma$} &\multicolumn{1}{c|}{0.95}	&\multicolumn{1}{c}{0.99}\\ 

\multicolumn{2}{c|}{Refresh factor $\zeta$ } &\multicolumn{1}{c|}{$10^{-2}$}	&\multicolumn{1}{c}{$10^{-2}$}\\ 

\multicolumn{2}{c|}{Batch size $\lVert \mathcal{D} \rVert$} &\multicolumn{1}{c|}{120}	&\multicolumn{1}{c}{1256}\\

\multicolumn{2}{c|}{Optimizer} &\multicolumn{1}{c|}{Adam}	&\multicolumn{1}{c}{Adam}\\
\hline
\end{tabular}
\label{table1}
\end{table}

\begin{figure}[t]
\centering
\subfigure[]{
\includegraphics[width=0.48\textwidth]{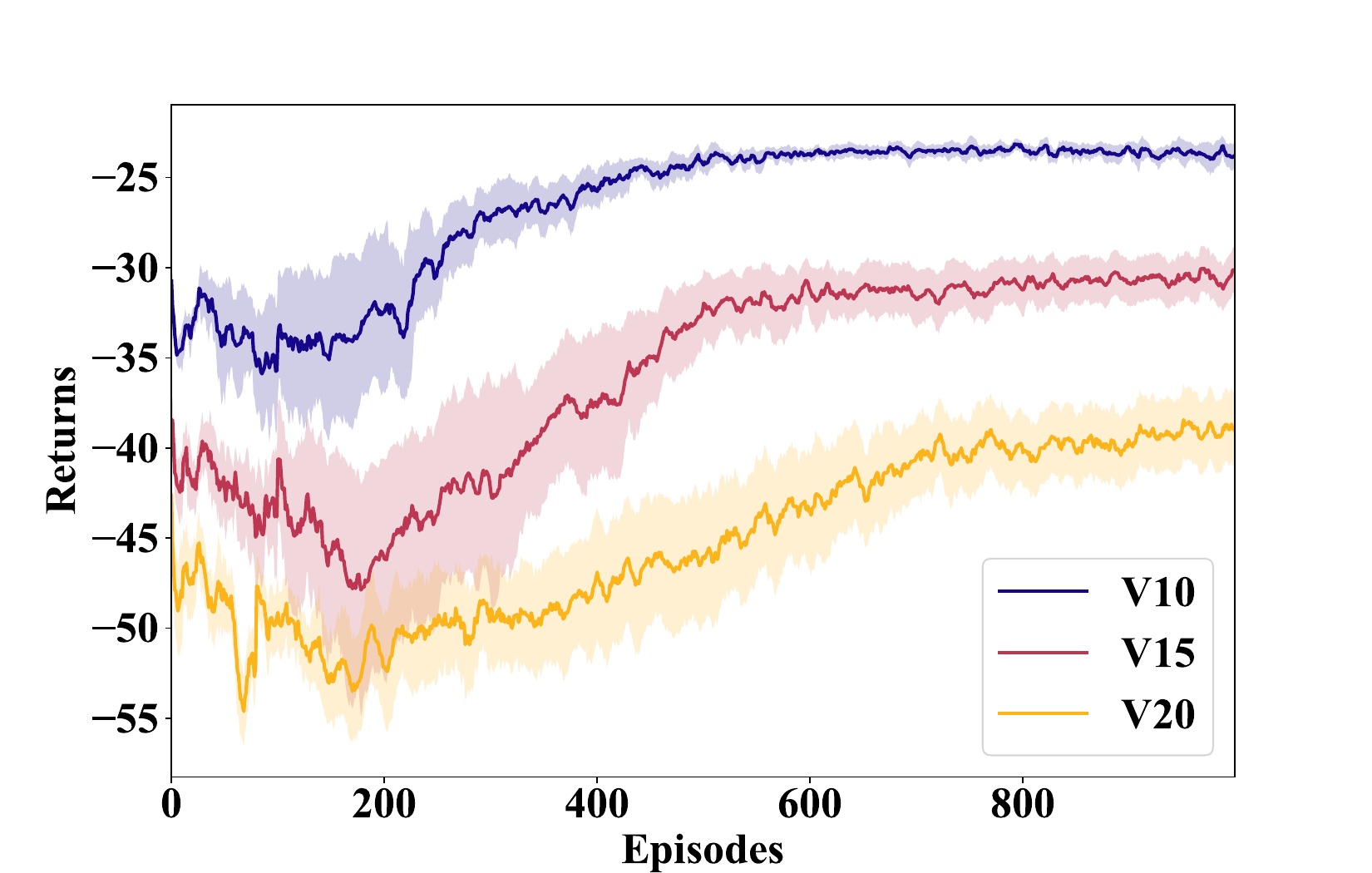} 
}
\hfil
\subfigure[]{
\includegraphics[width=0.48\textwidth]{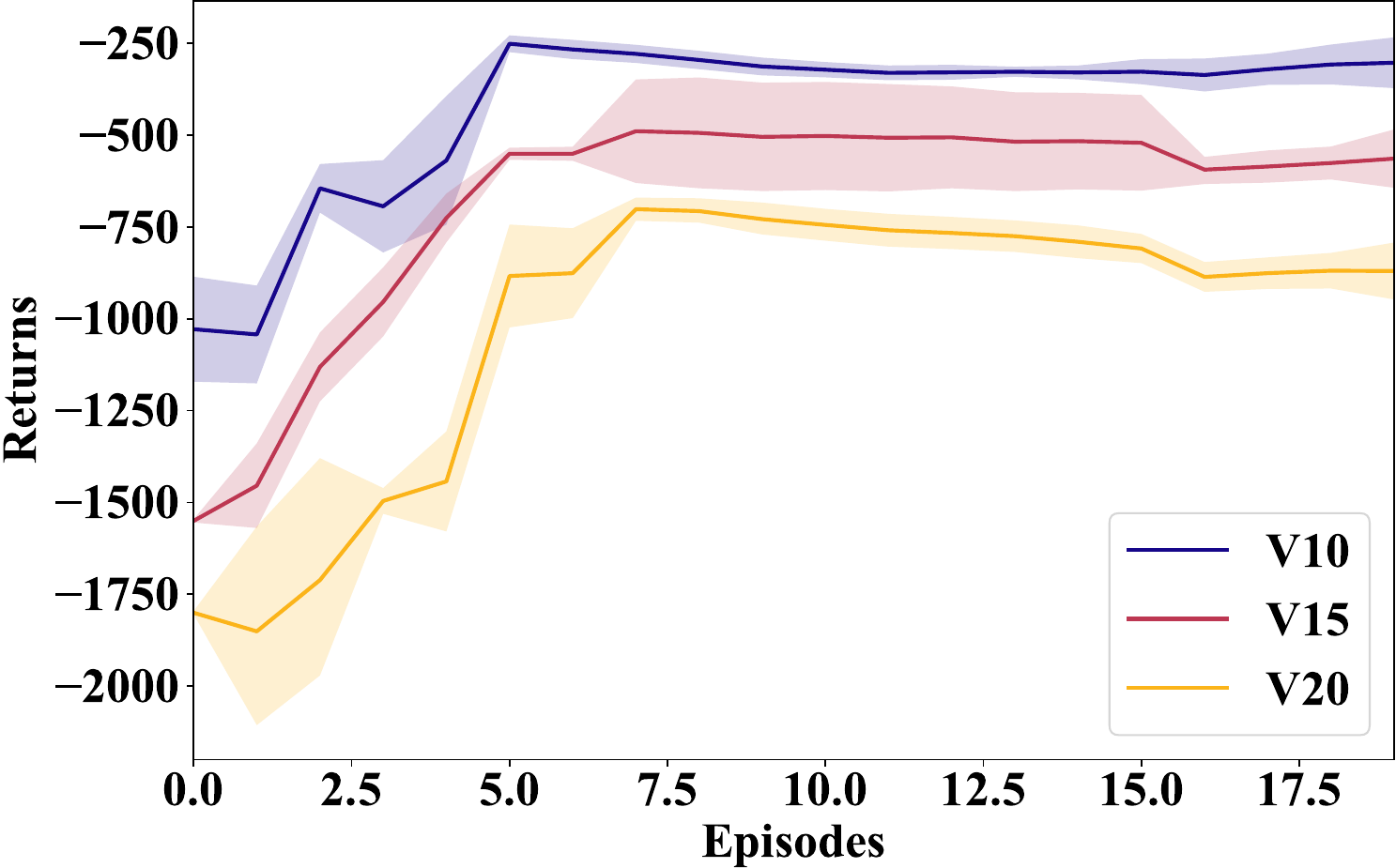} 
}
\hfil
\DeclareGraphicsExtensions.
\caption{Learning curves of the pre--training for upper layer and lower layer. (a) Target allocation. (b) Path planning.}
\label{fig4}
\end{figure}

\subsection{Setting Up}
\begin{figure*}[t]
    \centering
    \includegraphics[width=0.9\textwidth]{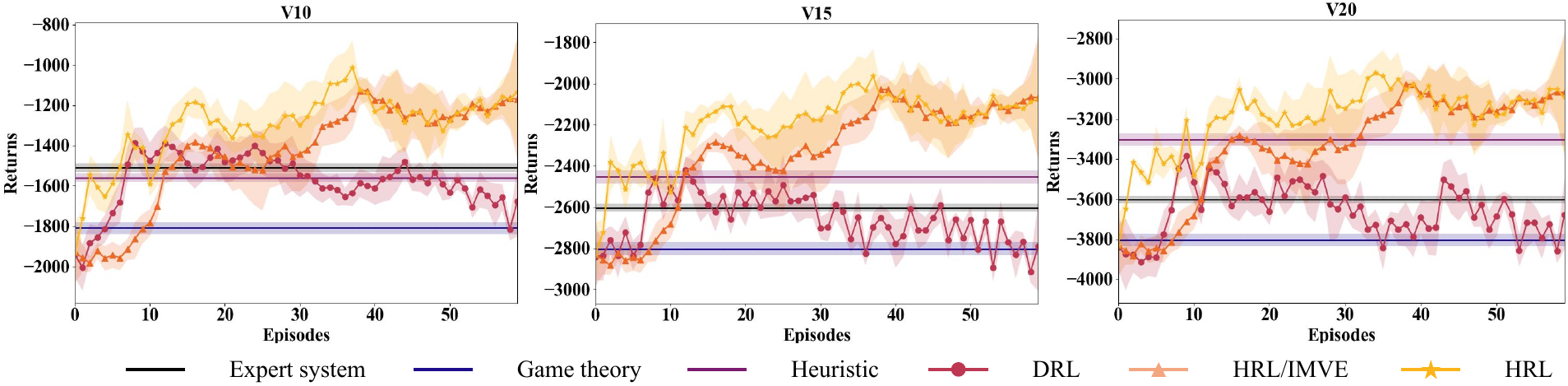}
    \caption{Learning curves of our method and baselines in different--size swarms.}
    \label{fig5}
\end{figure*}
\begin{table*}[t]
\renewcommand{\arraystretch}{1.25}
\setlength{\tabcolsep}{12pt}
\centering
\caption{Experiment Results of Our Method and Baselines in Different--Size Swarms.}
\label{table2}
\begin{tabular}{|c|c|c|c|c|c|c|c|c|c|c|}
\hline
\multicolumn{2}{c|}{\multirow{2}{*}{Method}} &\multicolumn{3}{c|}{V10} &\multicolumn{3}{c|}{V15} & \multicolumn{3}{c}{V20}\\

\multicolumn{2}{c|}{} & \multicolumn{1}{c}{Re.} & \multicolumn{1}{c}{Ti.(s)} & \multicolumn{1}{c|}{W.R.(\%)} & \multicolumn{1}{c}{Re.} & \multicolumn{1}{c}{Ti.(s)} & \multicolumn{1}{c|}{W.R.(\%)} & \multicolumn{1}{c}{Re.} & \multicolumn{1}{c}{Ti.(s)} & \multicolumn{1}{c}{W.R.(\%)}\\
\hline

\multicolumn{2}{c|}{Expert system} &\multicolumn{1}{c}{-1521}	&\multicolumn{1}{c}{8.26}	&\multicolumn{1}{c|}{86}	&\multicolumn{1}{c}{-2603}	&\multicolumn{1}{c}{8.89}	&\multicolumn{1}{c|}{79}	&\multicolumn{1}{c}{-3610}	&\multicolumn{1}{c}{9.67}	&\multicolumn{1}{c}{72}\\ 

\multicolumn{2}{c|}{Game theory} &\multicolumn{1}{c}{-1816}	&\multicolumn{1}{c}{4.39}	&\multicolumn{1}{c|}{69}	&\multicolumn{1}{c}{-2809}	&\multicolumn{1}{c}{7.71}	&\multicolumn{1}{c|}{72}	&\multicolumn{1}{c}{-3804}	&\multicolumn{1}{c}{10.43}	&\multicolumn{1}{c}{68}\\ 

\multicolumn{2}{c|}{Heuristic} &\multicolumn{1}{c}{-1564}	&\multicolumn{1}{c}{6.45}	&\multicolumn{1}{c|}{85}	&\multicolumn{1}{c}{-2435}	&\multicolumn{1}{c}{13.58}	&\multicolumn{1}{c|}{83}	&\multicolumn{1}{c}{-3302}	&\multicolumn{1}{c}{25.30}	&\multicolumn{1}{c}{80}\\

\multicolumn{2}{c|}{DRL} &\multicolumn{1}{c}{-1703}	&\multicolumn{1}{c}{\textbf{0.49}}	&\multicolumn{1}{c|}{77}	&\multicolumn{1}{c}{-2847}	&\multicolumn{1}{c}{\textbf{0.73}}	&\multicolumn{1}{c|}{70}	&\multicolumn{1}{c}{-3786}	&\multicolumn{1}{c}{\textbf{1.06}}	&\multicolumn{1}{c}{69}\\

\multicolumn{2}{c|}{UQ-HRL/IMVE} &\multicolumn{1}{c}{-1182}	&\multicolumn{1}{c}{0.86}	&\multicolumn{1}{c|}{93}	&\multicolumn{1}{c}{\textbf{-2091}}	&\multicolumn{1}{c}{1.20}	&\multicolumn{1}{c|}{90}	&\multicolumn{1}{c}{-3078}	&\multicolumn{1}{c}{1.87}	&\multicolumn{1}{c}{86}\\

\multicolumn{2}{c|}{UQ-HRL} &\multicolumn{1}{c}{\textbf{-1177}}	&\multicolumn{1}{c}{0.87}	&\multicolumn{1}{c|}{\textbf{94}}	&\multicolumn{1}{c}{-2093}	&\multicolumn{1}{c}{1.22}	&\multicolumn{1}{c|}{\textbf{90}}	&\multicolumn{1}{c}{\textbf{-3071}}	&\multicolumn{1}{c}{1.87}	&\multicolumn{1}{c}{\textbf{87}}\\
\hline
\end{tabular}
\label{table2}
\end{table*}

Before analyzing the comparative results, we first introduce the detailed settings of simulations. Following the convention in \cite{hou2023hierarchical,nian2024large,qu2023pursuit}, we generate the starting locations of pursuers and evaders in the rectangular area ($D_1 = 40m, D_2 = 20m, D_3 = 30m$). The target point of evaders is behind the pursuers. Three scenarios including ten pursuers versus ten evaders (termed as V$10$), fifteen pursuers versus fifteen evaders (termed as V$15$), and twenty pursuers versus twenty evaders (termed as V$20$) are considered in our experiment, all in the presence of four moving obstacles. In each scenario, we set up five different abilities of pursuers and evaders. The capture radius of pursuers $\rho_c$ is equalized from $0.6m$ to $1m$. The maximum velocity magnitudes of pursuers are all set to $0.5m/s$. The maximum velocity magnitudes of evaders are equalized from $0.6m/s$ to $1m/s$. The radius of the agents and obstacles are $0.2m$ and $0.5m$, respectively. Each obstacle keeps moving at a constant velocity over a certain time interval, while the magnitude $\lVert v_o \rVert \in [0.2, 0.5]$ and direction $\psi_o \in [-\pi, \pi]$ of the velocity change randomly between every interval.

For algorithm training, we pre--train the upper layer and lower layer for $E_{p,u}=300$ and $E_{p,l}=20$ episodes in $100$ randomly generated instances, respectively. The training steps of upper layer $T_u$ are equal to the number of pursuers, and the training steps of lower layer $T_l=300$. Then, the cross--training is adopted for $E_c = 60$ episodes in the generated instances above. In this way, we can enhance the training efficiency and stability of the proposed method. We list all the hyperparameters in Table \ref{table1} to demonstrate the details of our algorithm.

We use the artificial potential field method with attractive and repulsive forces to find a motion vector for evaders. The preset target point exerts an attractive force in the direction of the vector between the target point and the evader. In addition, the pursuers, neighbors, and obstacles exert repulsive forces so that the evader can avoid being captured or collision. The repulsive forces decrease proportionally to the distance squared. The velocity is denoted as 
\begin{equation}
    \begin{array}{c}
        \displaystyle v_j = \frac{p_{\text{tar}}-p_j}{\lVert p_{\text{tar}}-p_j \rVert} + \sum_{\substack{w\in \left[ U_p,U_e,O \right], \\ w \neq j}}{\frac{p_j-p_w}{\lVert p_j-p_w \rVert ^2}},
    \end{array}\label{equ38}
\end{equation}
where $p_{\text{tar}}$ is the position of the target point.

\subsection{Learning Performance in Pre--training}
We conduct these simulations on a server with a Windows 10 operating system, Intel Core i7--11700 CPU, 16--GB memory, and Radeon 520 GPU. All simulation programs are developed based on Python 3.7 and PyCharm 2022.2.3 compiler. The learning curves for each scenario of the upper layer and lower layer during the pre--training process are shown in Fig. \ref{fig4}. The horizontal axis refers to the number of episodes. The vertical axis of target allocation and path planning refers to the episode returns calculated by (\ref{equ4}) and (\ref{equ13}), respectively. To plot experimental curves, we adopt solid curves to depict the mean of all instances and shaded regions corresponding to standard deviation among instances. We can observe that the completely randomized experience replay can lead to performance fluctuations in the initial stage due to the use of the centralized algorithm in the upper layer. Since the rewards of target allocation and path planning are related to the number of agents, the episode returns gradually decrease as the problem size of swarm confrontation increases. The curves for both upper and lower converge stably in different sizes, suggesting that they have learned valid policies. Based on the pre--trained upper layer and lower layer, we can cross--train the two layers and compare our method with other baselines.

\subsection{Comparison Analysis in Cross--Training}

In the cross--training process, we employ the proposed method to improve the performance of the two layers. We compare the method with the baselines including the expert system, game theory, heuristic approach, and traditional DRL. IMVE adopts a rolling optimization approach to enhance the sample utilization of the upper layer. To investigate the influence of IMVE, we further conduct an ablation study with HRL absent on the IMVE method. The baselines are briefly introduced as follows.
\begin{itemize}
\item[1)] Expert system: Based on the rules of the target allocation and path planning in \cite{hou2023hierarchical}, the algorithm can match the optimal action to the current state. Therefore, it is also a rule--based swarm confrontation approach.
\item[2)] Game theory: The algorithm models the scenario as a differential game and seeks strategies using Nash equilibrium in a two--coalition non--cooperative game \cite{liu2022distributed}, which ultimately obtains the pursuers' strategies.
\item[3)] Heuristic algorithm: The algorithm constructs biologically inspired mobile adaptive networks to imitate the dynamics confrontation of swarms \cite{xia2023dynamic}. By building a distributed modular framework that includes target allocation and path planning, pursuers make successive and prompt decisions.
\item[4)] DRL: The algorithm considers swarm confrontation as an MDP process and directly employs multi--agent reinforcement learning to output the pursuers' strategies end--to--end \cite{qu2023pursuit}.
\item[5)] HRL/IMVE: An approach adopts the uncertainty quantification as our HRL method, while it updates the upper layer without IMVE.
\end{itemize}

The learning curves of HRL and baselines in different--size swarms are presented in Fig. \ref{fig5}. Expert system, game theory, and heuristic algorithms can develop strategies for pursuers in different instances. Among these three non--learning algorithms, game theory performs the worst in uncertainty scenarios. The heuristic algorithm outperforms the expert system algorithm in large--scale swarms. Learning--based algorithms do not perform as well as non--learning algorithms at first due to the randomness of the initial strategy. However, with continuous exploration and training, their performance gradually surpasses the latter. Due to the hybrid decision space in swarm confrontation, the performance of DRL decreases continuously in the later training. In contrast, HRL can find better strategies than non--learning algorithms in different sizes. In addition, with the IMVE method, the performance curve converges in fewer episodes, effectively improving the learning efficiency of HRL. 

We further deploy the policy networks trained by the learning--based methods in $100$ different instances and compare them with the non--learning algorithms. We propose two additional evaluation metrics: decision time and confrontation win rate. Decision time refers to the total time taken by all pursuers for target allocation and path planning based on their observations. We take the average value after running different instances. The confrontation win rate refers to the ratio of the number of pursuers' successes to the number of all instances. The average experiment results of $100$ instances in different sizes are shown in Table \ref{table2}. The table gathers the episode returns (Re.), decision time (Ti.), and confrontation win rate (W.R.) of all methods. We can observe that there is a relationship between the win rate and episode returns, which is calculated by the reward function in this study. Pursuers train their strategies to achieve higher returns, which leads to a greater win rate in the confrontation. Among these three non--learning algorithms, the heuristic algorithm has a longer decision time, although its returns and win rate are higher as the problem size increases. Due to the end--to--end approach, DRL has the shortest decision time, but its returns and win rate are much less than HRL. IMVE enhances the training efficiency of the method, but there is no additional improvement in its decision--making performance in the case of policy convergence. As a result, all three metrics for HRL and HRL/IMVE are similar in different scales. The results show that our method achieves better episode returns, decision time, and confrontation win rate than the baselines, especially on large--scale instances. The reason for this is that swarm confrontation in a dynamic obstacle environment has a high degree of uncertainty, and the baselines lack the ability to handle it. Our method optimizes the dynamic mechanism between the two layers based on the probabilistic ensemble model, which quantifies the uncertainty in the scenario.

\begin{figure*}[t]
    \centering
    \includegraphics[width=0.9\textwidth]{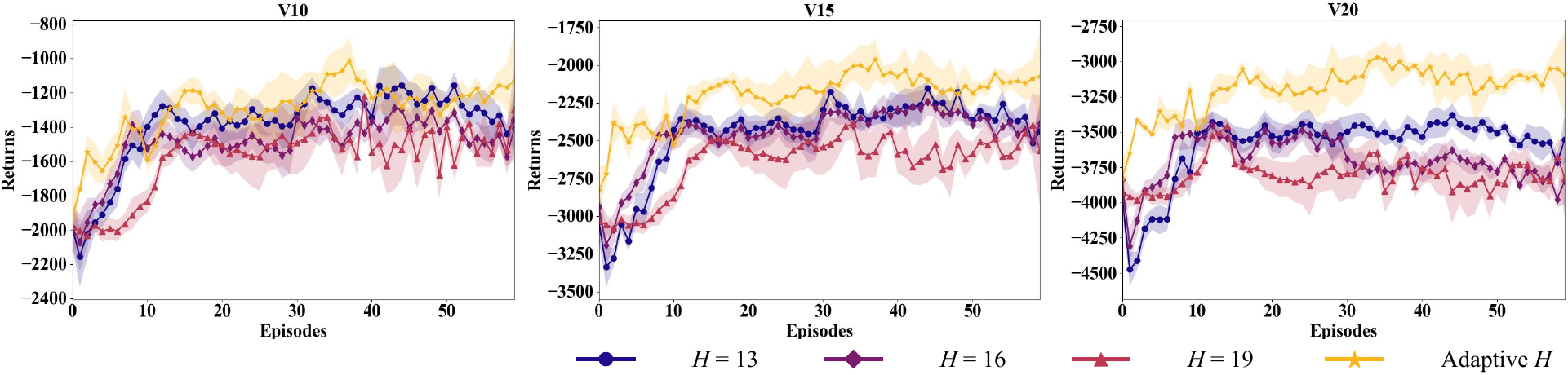}
    \caption{Learning curves of adaptive and fixed $H$ approach in different--size swarms.}
    \label{fig7}
\end{figure*}

\subsection{Ablation Study for Adaptive Interaction Frequency}
In our method, we construct a probabilistic ensemble model between the upper and lower to quantify the uncertainty. The method optimizes the interaction frequency between the two layers based on the adaptive truncation approach. With adaptive interaction frequency, we construct a dynamic mechanism between target allocation and path planning that enables pursuers to overcome uncertainty while chasing the evaders. To illustrate the effectiveness of the adaptive frequency approach in our method, we conduct an ablation study by fixing interaction step $H$ to three sets of constants ($H=13,16,19$). We compare our adaptive method with them in the cross--training process in different scenarios. The learning curves are shown in Fig. \ref{fig7}. When $H$ is a constant, the curves first decline for a while, and the decline is greater as $H$ is smaller, which means the target is assigned more frequently. This is due to the fact that the upper layer, which is only pre--trained, is less capable of handling uncertainty, and frequent use leads to a drop in performance. When the upper layer has gone through several episodes of cross--training, the curves will eventually converge to sub--optimal values, although they will rise rapidly. Furthermore, the error of sub--optimal values will become larger as the problem size increases.

\begin{figure*}[t]
    \centering
    \includegraphics[width=0.9\textwidth]{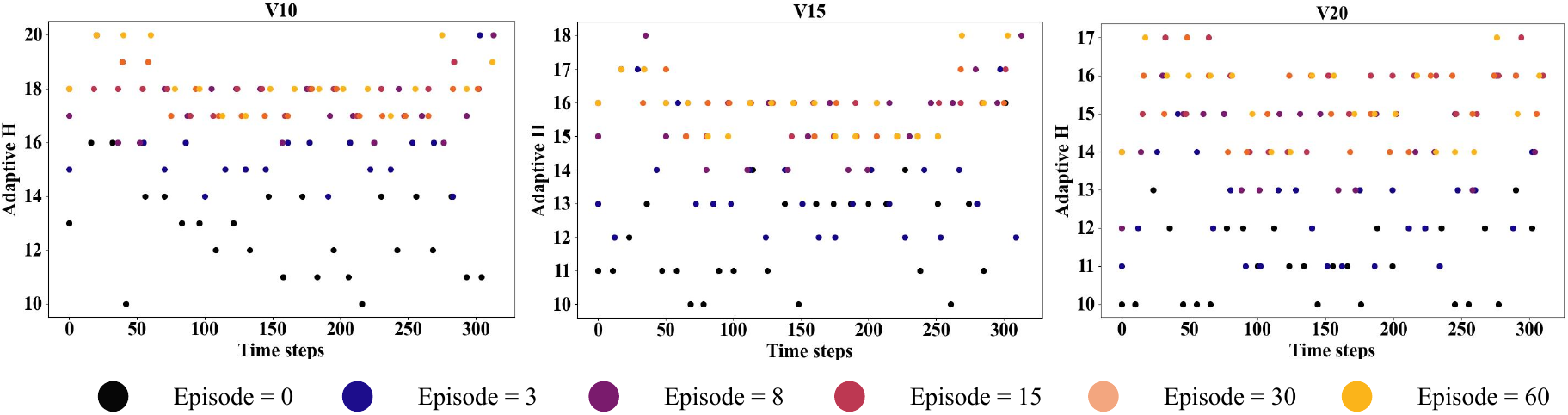}
    \caption{The value of $H$ for our method in different--size swarms.}
    \label{fig8}
\end{figure*}

\begin{figure*}[t]
    \centering
    \includegraphics[width=0.9\textwidth]{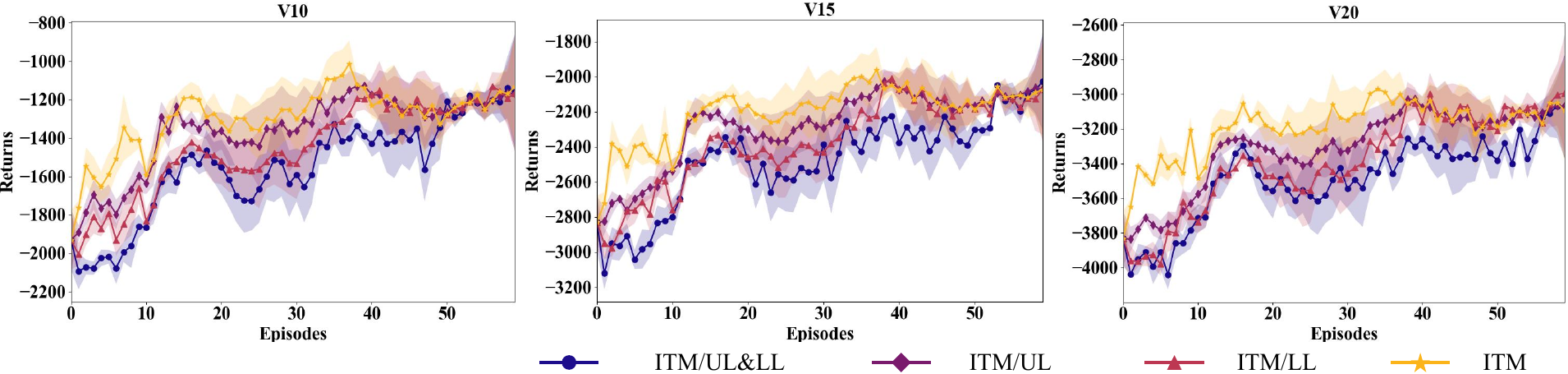}
    \caption{Ablation study results of ITM in different--size swarms.}
    \label{fig9}
\end{figure*}

Moreover, we plot the value of $H$ for the adaptive method in different sizes and different episodes ($0$--$60$) of cross--training in Fig. \ref{fig8}. The horizontal coordinate refers to the time steps in each episode, and the vertical coordinate refers to the value of $H$. In the early stage of training, the method decreases the interaction step $H$ to augment the training samples in the upper layer. Through constant training, the epistemic uncertainty in the lower layer decreases, so the interaction frequency becomes progressively lower. However, since aleatoric uncertainty cannot be eliminated, the increment in $H$ stabilizes after episode $15$. In addition, during time steps $50$--$250$, the interaction frequency will be higher than the other time steps because the pursuers will meet with evaders and obstacles, and the aleatoric uncertainty will be higher at this time. Based on the adaptive frequency method, we can overcome the negative effects of uncertainty on cross--training, and finally achieve a favorable training effect. 

\begin{table*}[t]
\renewcommand{\arraystretch}{1.25}
\setlength{\tabcolsep}{10pt}
\centering
\caption{Generalization Results of Our Method and Baselines in Different Size Swarms.}
\label{table3}
\begin{tabular}{|c|c|c|c|c|c|c|c|c|c|c|c|c|}
\hline
\multicolumn{2}{c|}{Number of} & \multicolumn{2}{c|}{\multirow{2}{*}{Method}} &\multicolumn{3}{c|}{V25} &\multicolumn{3}{c|}{V30} & \multicolumn{3}{c}{V35}\\

\multicolumn{2}{c|}{obstacles} & \multicolumn{2}{c|}{} & \multicolumn{1}{c}{Re.} & \multicolumn{1}{c}{Ti.(s)} & \multicolumn{1}{c|}{W.R.(\%)} & \multicolumn{1}{c}{Re.} & \multicolumn{1}{c}{Ti.(s)} & \multicolumn{1}{c|}{W.R.(\%)} & \multicolumn{1}{c}{Re.} & \multicolumn{1}{c}{Ti.(s)} & \multicolumn{1}{c}{W.R.(\%)}\\
\hline

\multicolumn{2}{c|}{\multirow{5}{*}{4}} & \multicolumn{2}{c|}{Expert system} & \multicolumn{1}{c}{-4738}	&\multicolumn{1}{c}{10.25}	&\multicolumn{1}{c|}{69}	&\multicolumn{1}{c}{-5763}	&\multicolumn{1}{c}{11.03}	&\multicolumn{1}{c|}{66}	&\multicolumn{1}{c}{-6984}	&\multicolumn{1}{c}{12.12}	&\multicolumn{1}{c}{62}\\ 

\multicolumn{2}{c|}{} & \multicolumn{2}{c|}{Game theory} &\multicolumn{1}{c}{-4799}	&\multicolumn{1}{c}{14.17}	&\multicolumn{1}{c|}{67}	&\multicolumn{1}{c}{-5905}	&\multicolumn{1}{c}{18.89}	&\multicolumn{1}{c|}{63}	&\multicolumn{1}{c}{-7149}	&\multicolumn{1}{c}{24.32}	&\multicolumn{1}{c}{60}\\ 

\multicolumn{2}{c|}{} & \multicolumn{2}{c|}{Heuristic} &\multicolumn{1}{c}{-4186}	&\multicolumn{1}{c}{40.82}	&\multicolumn{1}{c|}{78}	&\multicolumn{1}{c}{-5094}	&\multicolumn{1}{c}{61.34}	&\multicolumn{1}{c|}{76}	&\multicolumn{1}{c}{-6027}	&\multicolumn{1}{c}{88.73}	&\multicolumn{1}{c}{75}\\

\multicolumn{2}{c|}{} & \multicolumn{2}{c|}{DRL} &\multicolumn{1}{c}{-4832}	&\multicolumn{1}{c}{\textbf{1.58}}	&\multicolumn{1}{c|}{66}	&\multicolumn{1}{c}{-5871}	&\multicolumn{1}{c}{\textbf{2.10}}	&\multicolumn{1}{c|}{63}	&\multicolumn{1}{c}{-6758}	&\multicolumn{1}{c}{\textbf{2.86}}	&\multicolumn{1}{c}{65}\\

\multicolumn{2}{c|}{} & \multicolumn{2}{c|}{UQ-HRL} &\multicolumn{1}{c}{\textbf{-3911}}	&\multicolumn{1}{c}{2.41}	&\multicolumn{1}{c|}{\textbf{86}}	&\multicolumn{1}{c}{\textbf{-4862}}	&\multicolumn{1}{c}{3.05}	&\multicolumn{1}{c|}{\textbf{84}}	&\multicolumn{1}{c}{\textbf{-5703}}	&\multicolumn{1}{c}{4.20}	&\multicolumn{1}{c}{\textbf{83}}\\
\hline

\multicolumn{2}{c|}{\multirow{5}{*}{8}} & \multicolumn{2}{c|}{Expert system} & \multicolumn{1}{c}{-5392}	&\multicolumn{1}{c}{10.89}	&\multicolumn{1}{c|}{67}	&\multicolumn{1}{c}{-6499}	&\multicolumn{1}{c}{11.94}	&\multicolumn{1}{c|}{65}	&\multicolumn{1}{c}{-7841}	&\multicolumn{1}{c}{13.17}	&\multicolumn{1}{c}{60}\\ 

\multicolumn{2}{c|}{} & \multicolumn{2}{c|}{Game theory} &\multicolumn{1}{c}{-5521}	&\multicolumn{1}{c}{15.87}	&\multicolumn{1}{c|}{66}	&\multicolumn{1}{c}{-6737}	&\multicolumn{1}{c}{20.69}	&\multicolumn{1}{c|}{62}	&\multicolumn{1}{c}{-8120}	&\multicolumn{1}{c}{26.85}	&\multicolumn{1}{c}{57}\\ 

\multicolumn{2}{c|}{} & \multicolumn{2}{c|}{Heuristic} &\multicolumn{1}{c}{-4849}	&\multicolumn{1}{c}{48.95}	&\multicolumn{1}{c|}{80}	&\multicolumn{1}{c}{-5855}	&\multicolumn{1}{c}{70.47}	&\multicolumn{1}{c|}{75}	&\multicolumn{1}{c}{-6834}	&\multicolumn{1}{c}{99.51}	&\multicolumn{1}{c}{73}\\

\multicolumn{2}{c|}{} & \multicolumn{2}{c|}{DRL} &\multicolumn{1}{c}{-5296}	&\multicolumn{1}{c}{\textbf{1.74}}	&\multicolumn{1}{c|}{68}	&\multicolumn{1}{c}{-6344}	&\multicolumn{1}{c}{\textbf{2.59}}	&\multicolumn{1}{c|}{66}	&\multicolumn{1}{c}{-7295}	&\multicolumn{1}{c}{\textbf{3.37}}	&\multicolumn{1}{c}{65}\\

\multicolumn{2}{c|}{} & \multicolumn{2}{c|}{UQ-HRL} &\multicolumn{1}{c}{\textbf{-4194}}	&\multicolumn{1}{c}{2.88}	&\multicolumn{1}{c|}{\textbf{84}}	&\multicolumn{1}{c}{\textbf{-5213}}	&\multicolumn{1}{c}{3.61}	&\multicolumn{1}{c|}{\textbf{83}}	&\multicolumn{1}{c}{\textbf{-6159}}	&\multicolumn{1}{c}{4.93}	&\multicolumn{1}{c}{\textbf{80}}\\
\hline
\end{tabular}
\label{table3}
\end{table*}

\subsection{Ablation Study on ITM}
In ITM, we pre--train the upper layer and lower layer based on the rewards referred to (\ref{equ4}) and (\ref{equ13}), respectively. The pre--training phase allows the upper layer to learn a static allocation strategy and the lower layer to train an initial path planning policy. We conduct an ablation study in the cross--training to investigate the efficiency of pre--training for the upper and lower. The learning curves are shown in Fig. \ref{fig9}, where "ITM/X" refers to adopting ITM but without the pre--training in the "X". For example, ITM/UL denotes the training process only includes the pre--training in the lower layer and cross--training. Due to the lack of pre--training in the upper layer, the training efficiency of ITM/UL is significantly reduced. The curves of ITM/LL first decline for a while, which is attributed to the poor ability of the lower layer that lacks pre--training to deal with uncertainty. The absence of pre--training on two layers leads ITM/UL\&LL to perform the worst in cross-training, although it does not affect the final convergence value. Moreover, the effect of pre--training becomes more obvious as the problem size increases. The results verify that ITM can effectively speed up the learning process and avoid training instability.

\begin{figure}[t]
\centering
\subfigure[]{
\includegraphics[width=0.48\textwidth]{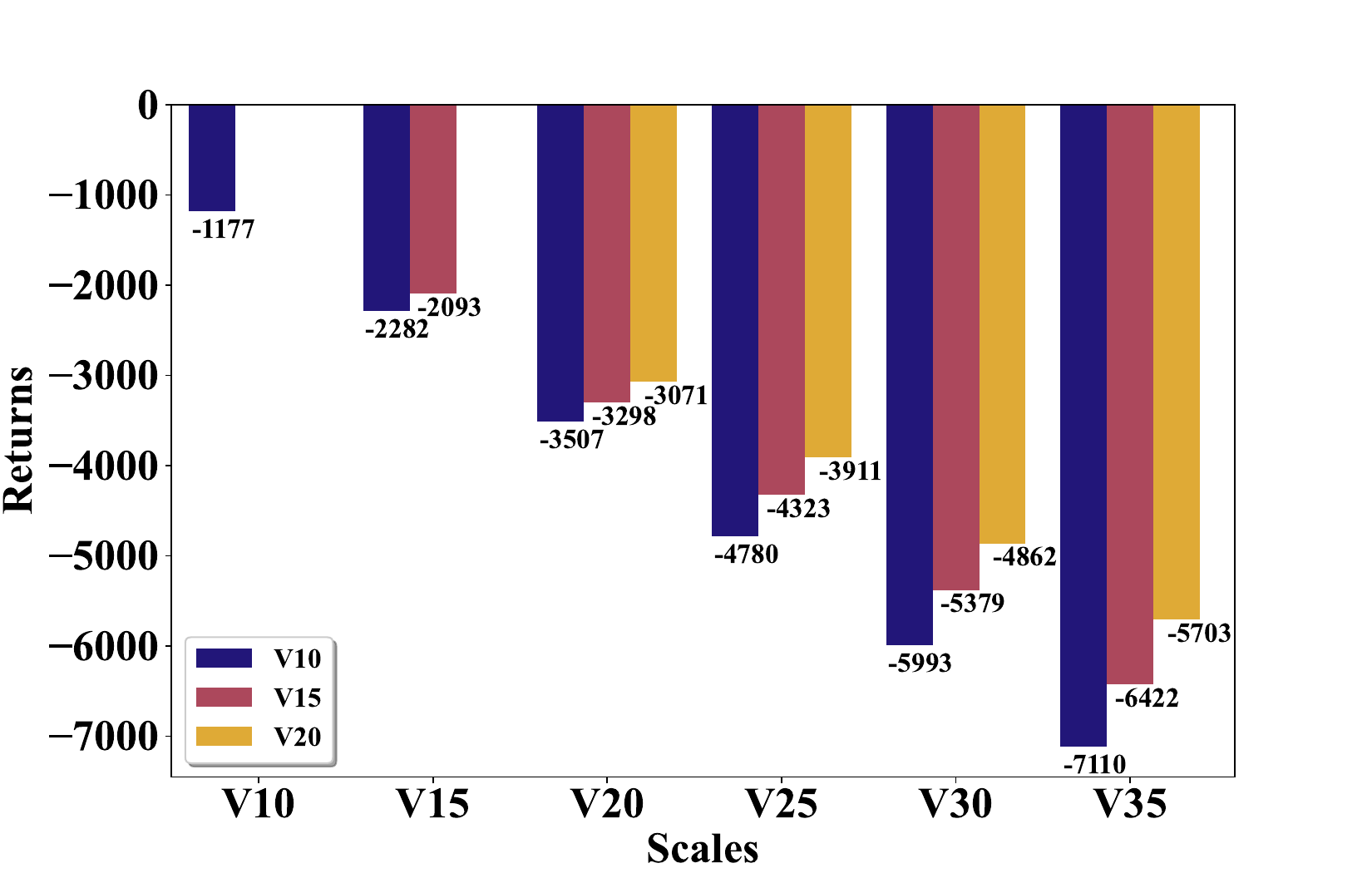} 
}
\hfil
\subfigure[]{
\includegraphics[width=0.48\textwidth]{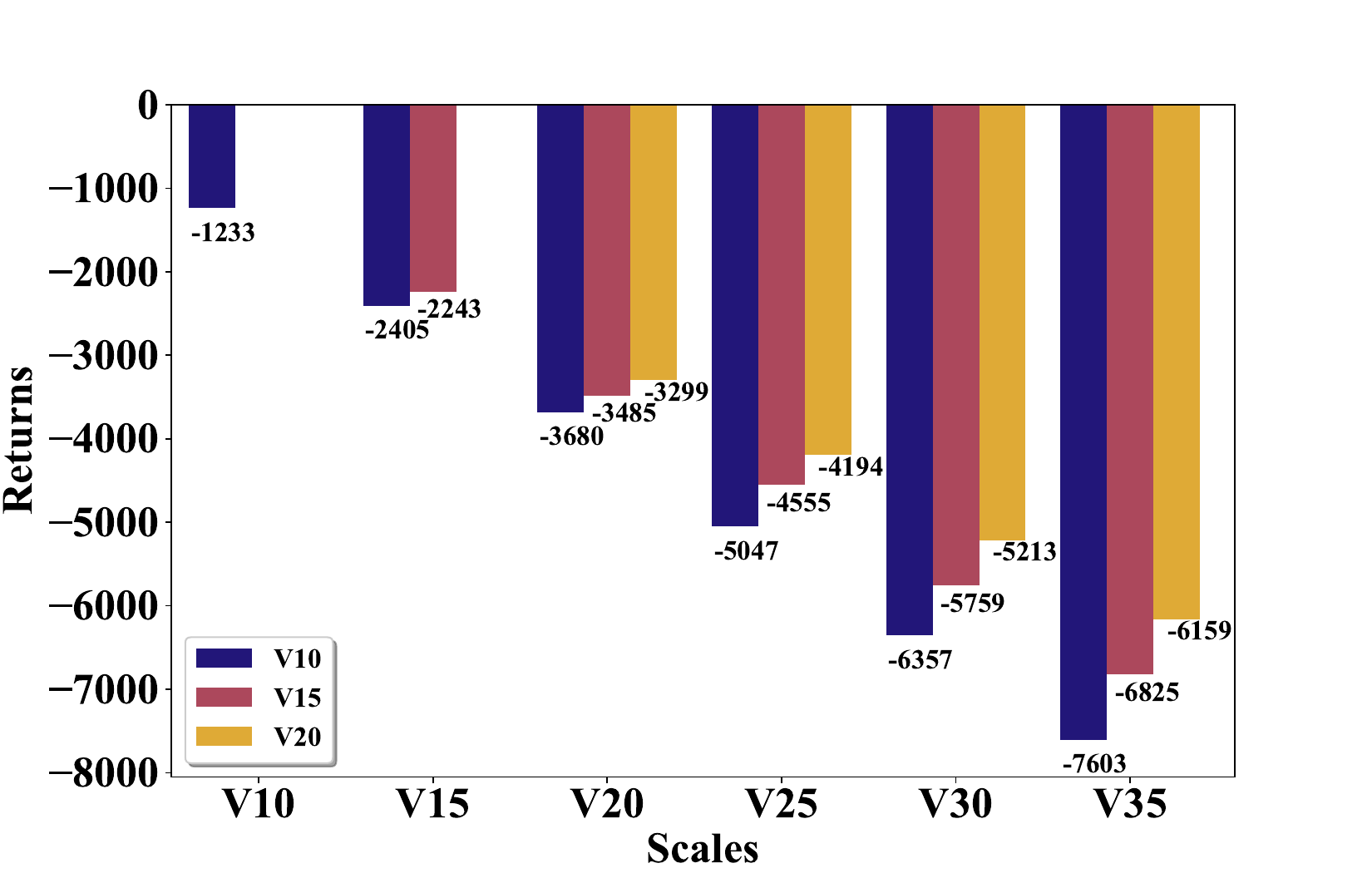} 
}
\hfil
\DeclareGraphicsExtensions.
\caption{Generalization to larger--scale instances. (a) Four obstacles. (b) Eight obstacles.}
\label{fig10}
\end{figure}

\subsection{Generalization to Larger--Size Swarms}
\begin{figure*}[t]
    \centering
    \includegraphics[width=0.9\textwidth]{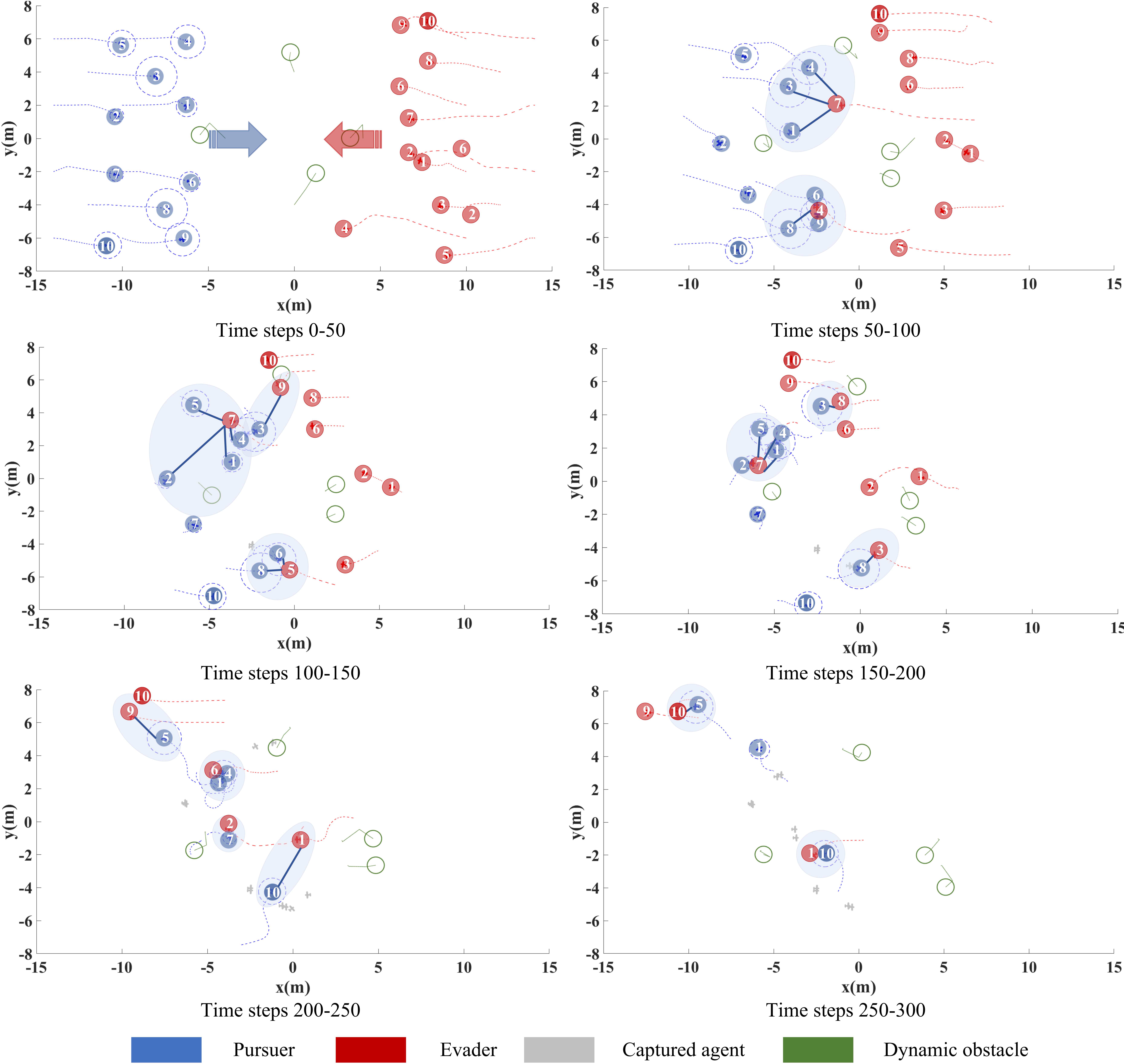}
    \caption{The swarm confrontation process with HRL in scenario V10.}
    \label{fig6}
\end{figure*}

In larger--scale swarm confrontation scenarios including V25, V30, and V35, we will lose a lot of computational resources by retraining the model. Therefore, we expect the trained model to have favorable generalization performance. We employ the trained two--layer networks to solve the instances with more agents and obstacles to verify the generalization performance of HRL. The models we trained in V10, V15, and V20 are generalized to solve the problem of larger sizes. The generalization results are shown in Fig. \ref{fig10}, where the horizontal axis refers to the scenario sizes, and the vertical axis refers to the episode returns. The legend refers to the model we trained in different sizes. Among scenarios V10--V20, the model trained for a certain size performs best on the corresponding scenario compared to those trained for other sizes. Although models trained on a smaller size have lower returns, they still outperform expert system, game theory, heuristic approach, and DRL. Due to the uncertainty associated with the increase in the number of obstacles, the path planning of the agents becomes more complex and the path reward is reduced. We observe that the model trained in V20 performs best on scenarios V25--V35, since it has a higher capability to handle uncertainty than the other models. 

We compare the model trained in V20 with the baselines in scenarios V25--V35 similar to Table \ref{table2}. The average experiment results are shown in Table \ref{table3}, which displays the episode returns (Re.), decision time (Ti.), and confrontation win rate (W.R.) testing in different problem sizes. As can be seen, HRL outperforms other baselines in all instances. To further investigate the generalization of our method, we utilize the trained model to solve instances with eight obstacles in Table \ref{table3}. An increase in the number of obstacles leads to a decrease in path rewards, but has a very small effect on confrontation win rate. The results demonstrate that the trained policies under small--size swarms are able to deal with the uncertainty in various--scale scenarios. Therefore, our method has a strong generalization on the swarm confrontation problem.

\subsection{Emerged Coordinated Confrontation Behaviors}
With sufficient training from scratch, the proposed HRL method can emerge effective target allocation and path planning strategies. Here, we visualize the results of trained networks with our method on one of the V10 instances in Fig. \ref{fig6}, and analyze the coordinated behaviors in each stage. Pursuers and evaders are labeled with "$X$" indicating their identification numbers. The figure illustrates the process of target allocation and path planning by pursuers through policy networks at different time steps. In the figure, the upper layer assigns targets based on pursuers' and evaders' attributes to maximize the total reward in (\ref{equ22}). The lower layer further plans the safe paths to chase evaders while avoiding collisions with obstacles and neighbors. 

Time steps $0$--$50$: The upper layer allows for an even target allocation of pursuers to ensure the best possible coverage of the defended area.

Time steps $50$--$100$: After time step $50$, the pursuers begin to encounter evaders and obstacles, introducing more environmental uncertainty into decision--making. Since pursuers with a large radius have a wider capture range, the upper layer will prioritize mobilizing them to chase evaders with faster escape velocities, such as evader $4$ and evader $7$.

Time steps $100$--$200$: In the round--up of evader $7$, the pursuers with a large capture radius surround evader $7$ by path planning, and then allow the pursuer $2$ with a smaller radius to conduct the final capture. This strategy preserves the pursuers with large radius by sacrificing the ones with smaller radius, which is beneficial to subsequent capture.

Time steps $200$--$250$: After the evaders have escaped the initial round--up by the pursuers, the latter will choose to turn around and chase. The upper layer will assign the pursuers priority to evaders who are closer because they have a higher probability of being captured. Evaders may be hindered by the obstacles or neighbors ahead, resulting in a loss of velocities, and the pursuers eventually catch up with them.

Time steps $250$--$300$: Even though evader $9$ is approaching the target area, the others have been captured by pursuers. According to the rules in Section \ref{sec:sample3A}, pursuers capture more than half of the evaders, which win the confrontation. Therefore, our trained models could deliver reasonably favorable solutions in the swarm confrontation.

\begin{figure}[t]
    \centering
    \includegraphics[width=0.45\textwidth]{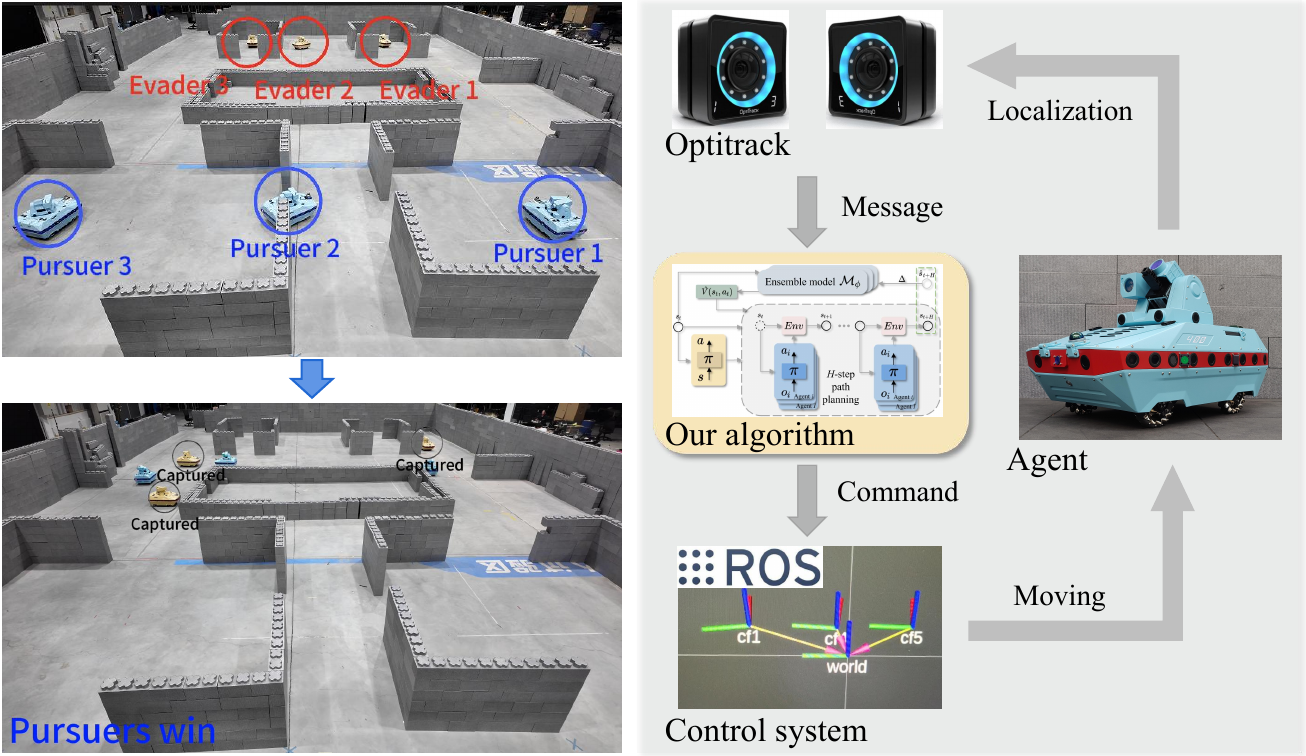}
    \caption{Deployment in the real robot system.}
    \label{fig11}
\end{figure}

\subsection{Deployment in Real--Robot System}
To verify the adaptability of our method in the real world, we conduct experiments with an actual robot system shown in Fig. \ref{fig11}. We consider three pursuers versus three evaders in a static--obstacle scenario, where its setting is consistent with the definition in Section \ref{sec:sample3A}. In the experiment, we maneuver agents by the ground control system with motion capture from Optitrack. By performing integration training in simulations, we can get the target allocation and path planning networks of pursuers, which can be deployed directly in real robots. Our algorithm receives messages from Optitrack, including the positions of all agents. Then the algorithm calculates the observation of each agent and outputs the position command with the above two networks, which is sent to the ground control system. Based on these networks, pursuers can capture all evaders while avoiding collision. The experiment's success verifies that the designed method can accomplish the pursuit--evasion game through offline training and online decision--making.

\section{Conclusion}
\label{sec:sample6}
This paper has presented a guaranteed stable HRL method for hybrid decision spaces and high uncertainty in swarm confrontation. It has designed two--layer DRL networks to reflect commands and actions with target allocation and path planning, and quantifies the uncertainty by constructing a probabilistic ensemble model. Furthermore, the method has optimized the interaction mechanism and enhanced the sample utilization based on the uncertainty quantification. We have also proposed an integration training method including pre--training and cross--training to improve the training efficiency and stability. The experiment results have shown that our method achieves better episode returns, decision time, and confrontation win rate than the baselines, especially on large--size swarms. The influence of the adaptive frequency approach and ITM has been verified via ablation studies. Moreover, we have demonstrated the generalization of our method by applying a model trained under small--size swarms to the larger ones. After sufficient training, our method could emerge effective swarm confrontation strategies for agents and be deployed directly in the real--robot system.

\bibliographystyle{IEEEtran}
\bibliography{trans}

\begin{IEEEbiography}[{\includegraphics[width=1in,height=1.25in,clip,keepaspectratio]{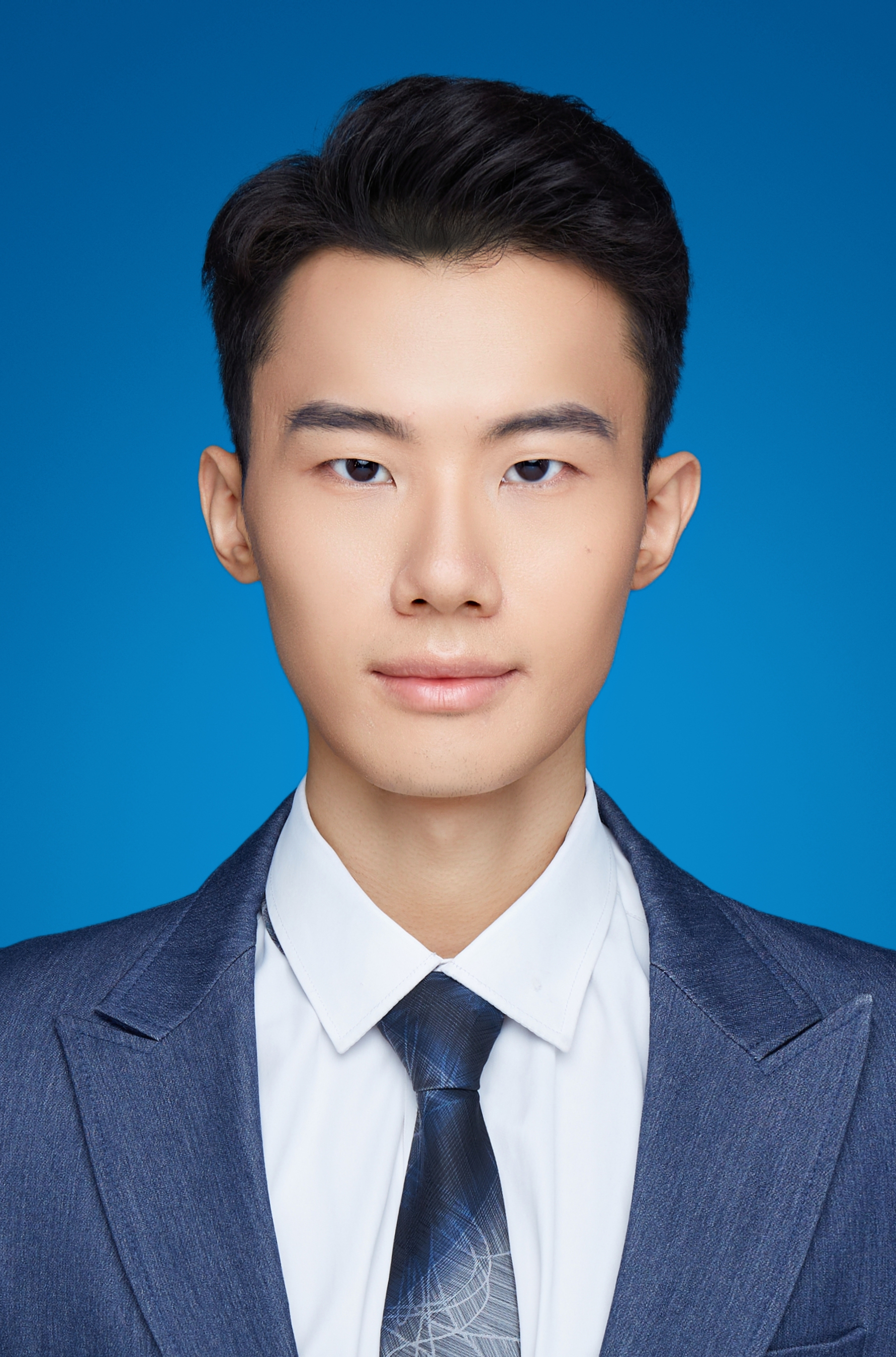}}]{Qizhen Wu} received the B.S. degree in aeronautical and astronautical engineering from Sun Yat--sen University, Guangzhou, China, in 2022. He is currently pursuing the Ph.D. degree with the School of Automation Science and Electrical Engineering, Beihang University, Beijing, China. His current research interests include reinforcement learning, robotic control, and swarm confrontation.
\end{IEEEbiography}
\begin{IEEEbiography}[{\includegraphics[width=1in,height=1.25in,clip,keepaspectratio]{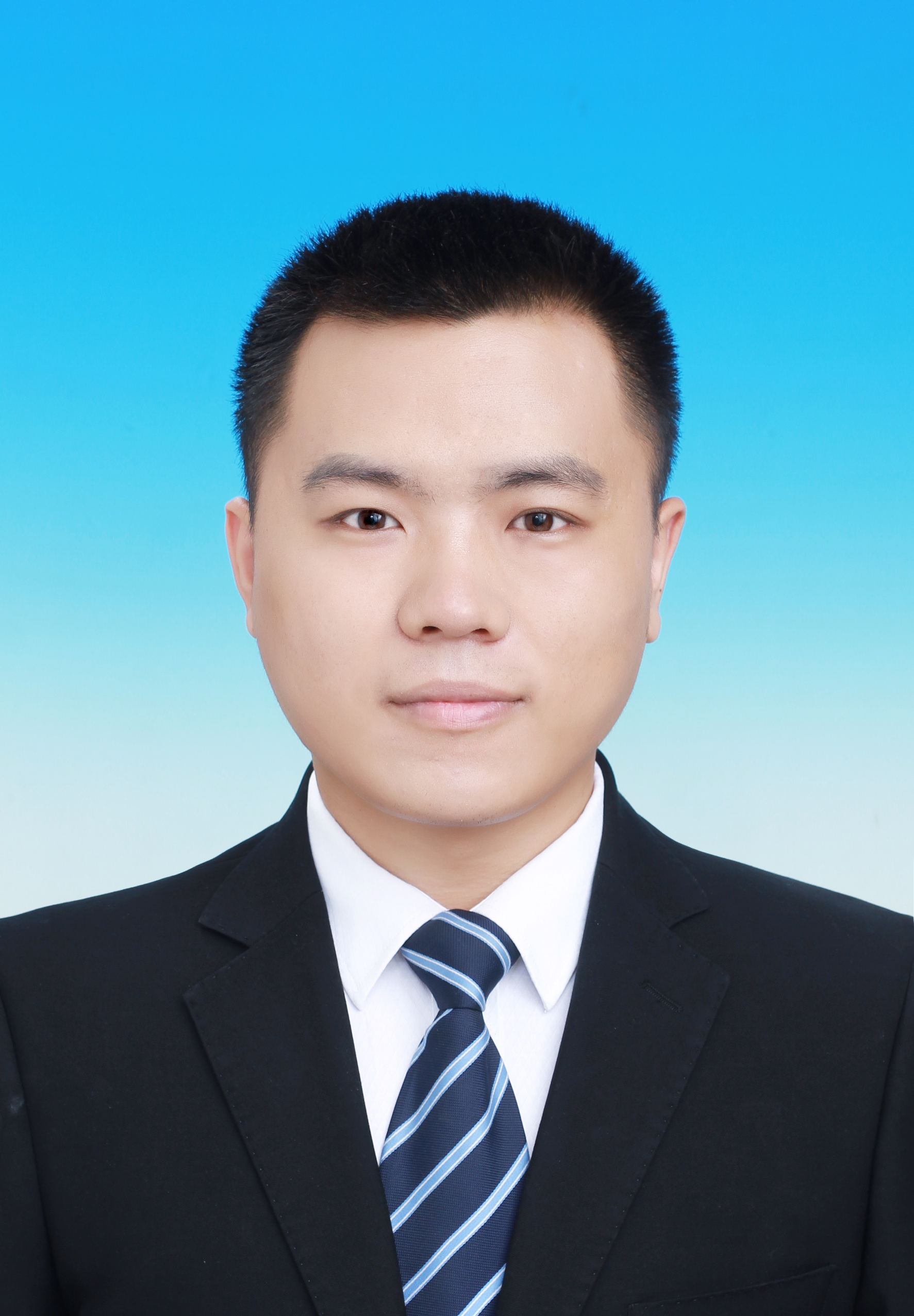}}]{Kexin Liu}
received the M.Sc. degree in control
science and engineering from Shandong University,
Jinan, China, in 2013, and the Ph.D. degree in
system theory from the Academy of Mathematics
and Systems Science, Chinese Academy of Sciences,
Beijing, China, in 2016.
From 2016 to 2018, he was a Postdoctoral
Fellow with Peking University, Beijing. He is currently an Associated Professor with the School
of Automation Science and Electrical Engineering,
Beihang University, Beijing. His current research interests include multi--agent systems and complex networks.
\end{IEEEbiography}
\begin{IEEEbiography}[{\includegraphics[width=1in,height=1.25in,clip,keepaspectratio]{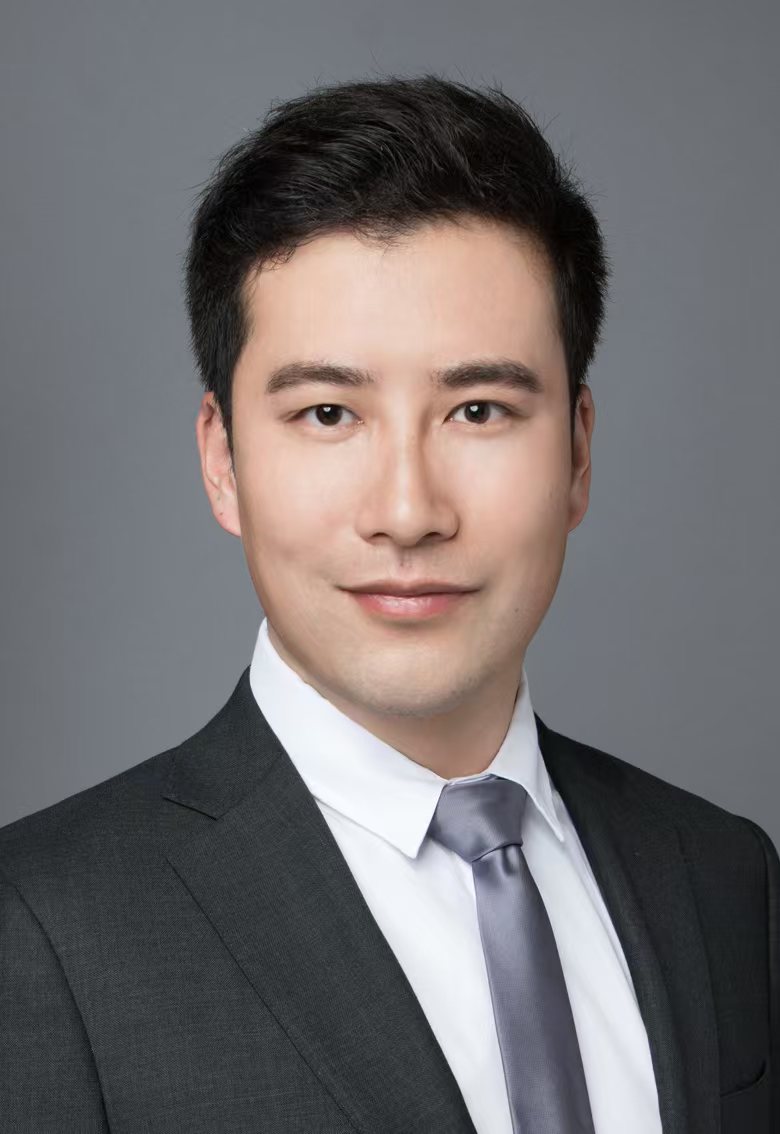}}]{Lei Chen}
received the Ph.D. degree in control
theory and engineering from Southeast University,
Nanjing, China, in 2018.
He was a visiting Ph.D. student with the Royal
Melbourne Institute of Technology University,
Melbourne, VIC, Australia, and Okayama
Prefectural University, Soja, Japan. From 2018
to 2020, he was a Postdoctoral Fellow with
the School of Automation Science and Electrical
Engineering, Beihang University, Beijing, China. He
is currently with the Advanced Research Institute
of Multidisciplinary Science, Beijing Institute of Technology, Beijing, as an
Associate Research Fellow. His current research interests include complex
networks, characteristic model, spacecraft control, and network control.
\end{IEEEbiography}
\begin{IEEEbiography}[{\includegraphics[width=1in,height=1.25in,clip,keepaspectratio]{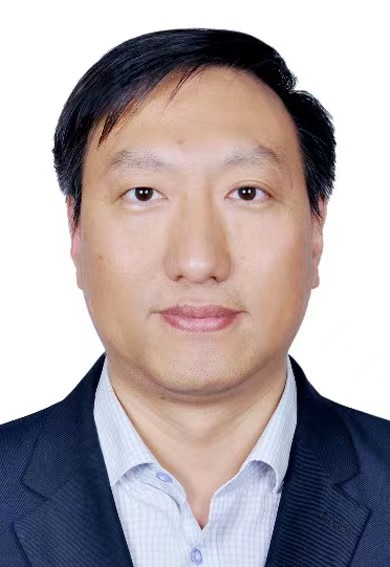}}]{Jinhu L\"u}
(Fellow, IEEE) received the Ph.D.
degree in applied mathematics from the Academy
of Mathematics and Systems Science, Chinese
Academy of Sciences, Beijing, China, in 2002.
He was a Professor with RMIT University,
Melbourne, VIC, Australia, and a Visiting Fellow
with Princeton University, Princeton, NJ, USA.
He is currently the Vice President of Beihang University, Beijing, China. He is also a Professor
with the Academy of Mathematics and Systems
Science, Chinese Academy of Sciences. He is a Chief Scientist of the National
Key Research and Development Program of China and a Leading Scientist
of Innovative Research Groups of the National Natural Science Foundation
of China. His current research interests include complex networks, industrial
Internet, network dynamics, and cooperation control.

Prof. L\"u was a recipient of the Prestigious Ho Leung Ho Lee Foundation
Award in 2015, the National Innovation Competition Award in 2020, the
State Natural Science Award three times from the Chinese Government
in 2008, 2012, and 2016, respectively, the Australian Research Council
Future Fellowships Award in 2009, the National Natural Science Fund for
Distinguished Young Scholars Award, and the Highly Cited Researcher Award
in engineering from 2014 to 2020. He is/was an Editor in various ranks for
15 SCI journals, including the Co--Editor--in--Chief of IEEE TRANSACTIONS
ON INDUSTRIAL INFORMATICS. He served as a member in the Fellows
Evaluating Committee of the IEEE CASS, the IEEE CIS, and the IEEE IES.
He was the General Co--Chair of IECON 2017. He is a Fellow of CAA.
\end{IEEEbiography}
\end{document}